\documentclass[journal]{IEEEtran}
\usepackage[utf8]{inputenc}

\usepackage{cite}

\usepackage[colorlinks=true, linkcolor=blue, citecolor=blue, urlcolor=blue]{hyperref}

\usepackage{url}            
\usepackage{microtype}      
\usepackage{lipsum}         

\usepackage{amsmath}
\usepackage{amsfonts}       
\usepackage{amssymb}        
\usepackage{bm}             
\usepackage{nicefrac}       
\usepackage{bbm}            
\usepackage{siunitx}        

\usepackage{graphicx}
\usepackage{xcolor}         
\usepackage[labelfont=bf]{caption} 
\usepackage{subcaption}     
\usepackage{float}
\usepackage{wrapfig}

\usepackage{booktabs}       
\usepackage{multirow}       
\usepackage{longtable}      
\usepackage{makecell}       
\usepackage{nicematrix}     

\usepackage{tikz}
\usepackage{pgffor} 
\usepackage[edges]{forest}
\usepackage{tikz-dependency}
\usepackage{bookmark}

\begin{document}
%
\title{Empowering Real-World: A Survey on the Technology, Practice, and Evaluation of LLM-driven Industry Agents}

%
%
%

\author{Yihong Tang,~\IEEEmembership{}
        Kehai Chen,~\IEEEmembership{}
        Liang Yue,~\IEEEmembership{}
        Jinxin Fan,~\IEEEmembership{}
        Caishen Zhou,~\IEEEmembership{}
        Xiaoguang Li, Yuyang Zhang, Mingming Zhao, Shixiong Kai, Kaiyang Guo, Xingshan Zeng, Wenjing Cun, Lifeng Shang, Min Zhang
\thanks{Yihong Tang, Kehai Chen, Liang Yue, Jinxin Fan, Caishen Zhou and Min Zhang are with the School of Computer Science and Technology, 
Harbin Institute of Technology, Shenzhen, China, 518055 
e-mail: (chenkehai@hit.edu.cn).} 
\thanks{Xiaoguang Li, Yuyang Zhang, Mingming Zhao, Shixiong Kai, Kaiyang Guo, Xingshan Zeng, Wenjing Cun and Lifeng Shang are with Huawei Technologies Co., Ltd.}
}

%
%

\markboth{Journal of \LaTeX\ Class Files,~Vol.~14, No.~8, August~2015}%
{Shell \MakeLowercase{\textit{et al.}}: Bare Demo of IEEEtran.cls for IEEE Journals}
%



\maketitle

\begin{abstract}
With the rise of large language models (LLMs), LLM agents capable of autonomous reasoning, planning, and executing complex tasks have become a frontier in artificial intelligence. However, how to translate the research on general agents into productivity that drives industry transformations remains a significant challenge. To address this, this paper systematically reviews the technologies, applications, and evaluation methods of industry agents based on LLMs. Using an industry agent capability maturity framework, it outlines the evolution of agents in industry applications, from "process execution systems" to "adaptive social systems." First, we examine the three key technological pillars that support the advancement of agent capabilities: Memory, Planning, and Tool Use. We discuss how these technologies evolve from supporting simple tasks in their early forms to enabling complex autonomous systems and collective intelligence in more advanced forms. Then, we provide an overview of the application of industry agents in real-world domains such as digital engineering, scientific discovery, embodied intelligence, collaborative business execution, and complex system simulation. Additionally, this paper reviews the evaluation benchmarks and methods for both fundamental and specialized capabilities, identifying the challenges existing evaluation systems face regarding authenticity, safety, and industry specificity. Finally, we focus on the practical challenges faced by industry agents, exploring their capability boundaries, developmental potential, and governance issues in various scenarios, while providing insights into future directions. By combining technological evolution with industry practices, this review aims to clarify the current state and offer a clear roadmap and theoretical foundation for understanding and building the next generation of industry agents.
\end{abstract}


%
\IEEEpeerreviewmaketitle

\section{Introduction}
%
%
%
%

\IEEEPARstart{I}{n}
recent years, large language models (LLMs) have made groundbreaking progress. Through pre-training on vast amounts of data, they exhibit unprecedented language understanding, generation, and reasoning capabilities \cite{openai2024gpt4technicalreport, chowdhery2022palmscalinglanguagemodeling, touvron2023llamaopenefficientfoundation}. However, LLMs, as static and state-less predictive models, are mainly limited to processing text input and generating corresponding outputs. They struggle to actively interact with the external world or perform complex tasks that require long-term memory and multi-step operations \cite{ye2025taskmemoryenginetme, tan2025prospectretrospectreflectivememory}. To overcome this limitation, researchers have used LLMs as the "brain" to build autonomous agents that can perceive the environment, plan, act, and learn from interactions \cite{Wang_2024}. These LLM-driven agents integrate memory modules, planning algorithms, and tool invocation interfaces, combining the cognitive abilities of LLMs with dynamic interactions in the environment, thus forming the prototype of general agents capable of autonomously achieving open-ended goals.

As general agents transition from theory to practice, their application scenarios inevitably shift from simple, general digital environments to complex, knowledge-intensive, and high-risk industry domains \cite{raza2025industrial}. This gives rise to the concept of "Industry Agents." Industry agents refer to autonomous or semi-autonomous systems deployed in specific business contexts that leverage domain knowledge and specialized tools to solve real-world industry problems. For example, Xia et al. demonstrate how LLM agents can orchestrate modular production systems by planning tasks, invoking low-level control interfaces, and interfacing with digital twins \cite{Xia_2023}. Compared to general-purpose agents, industry agents face more severe challenges. They must not only possess general cognitive abilities but also address industry-specific requirements, such as the high time sensitivity and risks in finance \cite{chen2025standardbenchmarksfail}, the authoritative knowledge and security compliance in healthcare \cite{wang2025surveyllmbasedagentsmedicine}, and the physical constraints and process complexity in manufacturing \cite{li2024largelanguagemodelsmanufacturing, Garcia2024FrameworkFL}. The key issue becomes how to integrate general agent frameworks with deep industry expertise, complex business processes, and stringent safety standards, thus transforming the potential of agents into real-world productivity.

Meanwhile, with the rapid development of LLM-based agent research, numerous excellent review papers have emerged, offering valuable perspectives from different dimensions to help us understand the field. Some reviews focus on the core technical modules of intelligent agents. For example,  \cite{zhang2024surveymemorymechanismlarge} systematically reviews the memory mechanisms of agents;  \cite{huang2024understandingplanningllmagents} classifies and analyzes planning capabilities; and  \cite{Qu_2025} provides a comprehensive overview of tool learning paradigms and implementations. Additionally, \cite{mei2025surveycontextengineeringlarge} optimizes the information load in the LLM reasoning process from the perspective of context engineering, offering important support for efficient agent interactions. These works lay the foundation for a deeper understanding of the technical details of agents. Other reviews focus on general agent architectures and capabilities. \cite{Wang_2024, masterman2024landscapeemergingaiagent} propose general agent frameworks and classify existing architectures. At the same time, works like  \cite{ke2025surveyfrontiersllmreasoning, gao2025surveyselfevolvingagentspath} explore the implementation paths for advanced capabilities such as reasoning and self-evolution. Notably, \cite{liu2025advanceschallengesfoundationagents} presents a modular, brain-inspired view of agent cognitive, perception, and operation modules, while also addressing key topics such as self-enhanced evolution, multi-agent systems, and secure deployment. Additionally, some reviews focus on specific application areas or advanced paradigms. For instance, reviews like  \cite{gridach2025agenticaiscientificdiscovery, ding2024largelanguagemodelagent} delve into the applications of agents in scientific discovery and financial trading. Meanwhile,  \cite{chen2025surveyllmbasedmultiagentsystem, singh2025agenticretrievalaugmentedgenerationsurvey} explore multi-agent systems and the Agentic RAG paradigm. Moreover, \cite{gao2023largelanguagemodelsempowered} provides a comprehensive review of LLM-enabled agent-based modeling and simulation, covering applications in information, physics, social, and hybrid scenarios. \cite{10.1145/3719341} focuses on autonomous research agents, proposing a systematic methodology and evaluation blueprint for their construction. Finally, \cite{hu2025surveyscientificlargelanguage} offers a data-centric system review and roadmap for the development of scientific LLMs and agents from the perspective of data and model co-evolution.

Despite these outstanding contributions, there remains a gap in providing a systematic framework that combines technological evolution, application practice, and capability levels, with a focus on industry implementation. To fill this gap, this paper provides a comprehensive review of LLM-based industry agents. Specifically, the review is organized into three main areas: the technological foundations, practical applications, and real-world evaluations of industry agents. First, we delve into the three core technologies supporting agent capabilities: memory, planning, and tool use, and discuss their technological evolution. Then, we present a panoramic view of industry agent applications across various sectors using a five-level maturity framework. Next, we systematically examine evaluation benchmarks and methods for both foundational and specialized industry capabilities, highlighting their limitations. Finally, we focus on the deep challenges faced by industry agents in practice, exploring their bottlenecks, future development, and strategies to address these challenges.

In summary, the contributions of this paper include:

Proposing a Capability Maturity Framework: We introduce an innovative industry agent capability maturity framework that provides a clear metric for assessing and understanding the role and value of agents in various industries.

Linking Technology and Application: We connect the evolution of the three core technologies—memory, planning, and tool use—with capability levels, showing how technological advances drive the progression of application practices.

Focusing on Industry Practices and Evaluations: We systematically review agent applications in key industries and professional evaluation benchmarks, aligning closely with real-world industrial needs and challenges.

With this unique perspective, we aim to bridge the gap between agent applications across various domains, contributing to the maturation and prosperity of agent in the real world.

\begin{figure*}[htbp]
  \centering
  \includegraphics[width=.95\linewidth]{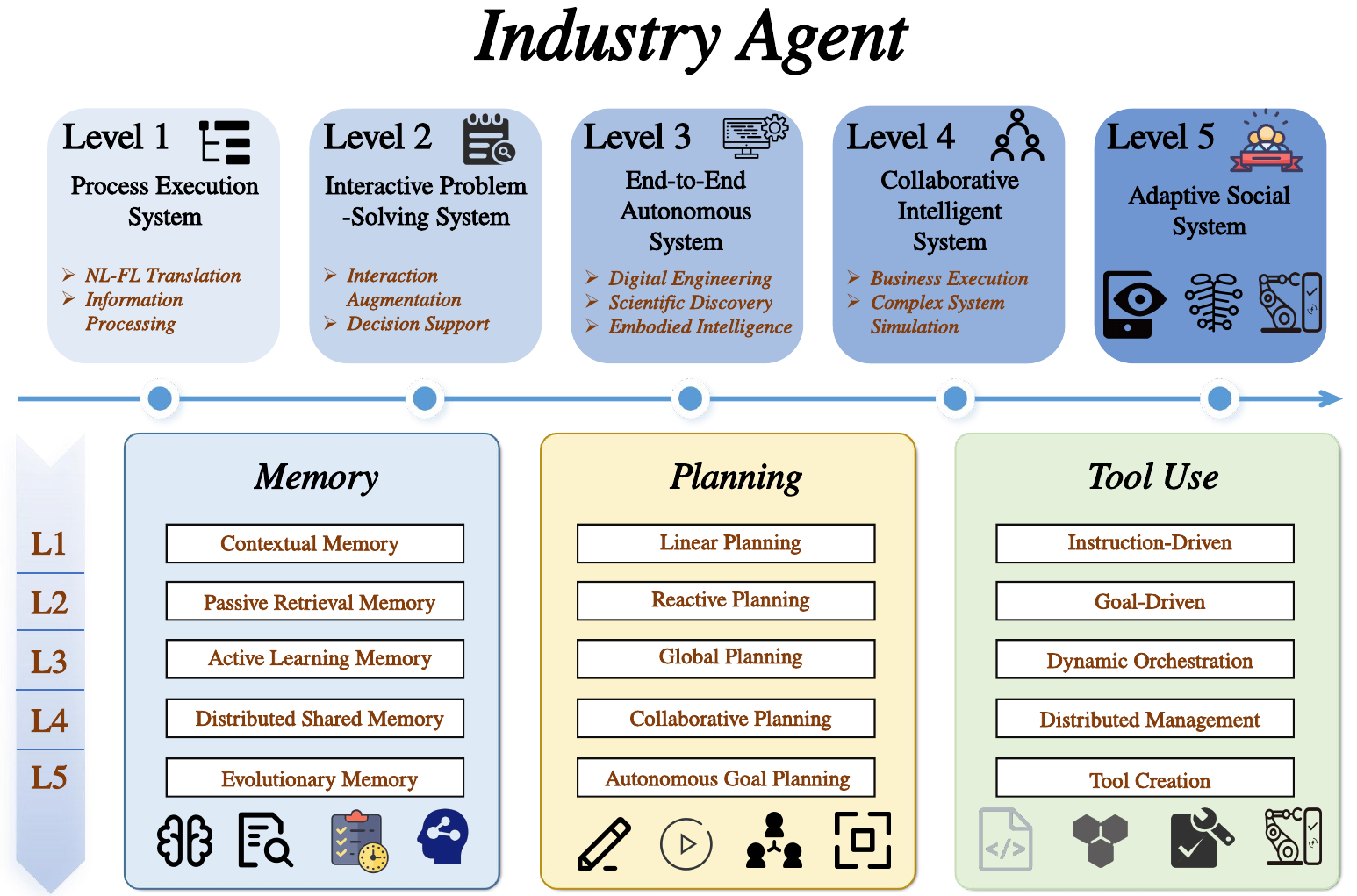}
  \caption{The framework of industry agent.}
  \label{fig: main}
  \end{figure*}

\tikzstyle{my-box}=[
    rectangle,
    draw=gray!50,
    rounded corners,
    text opacity=1,
    minimum height=2em, 
    minimum width=5em,
    inner sep=3pt,
    inner ysep=6pt, 
    align=center,
    fill opacity=0.15,
    line width=0.5pt,
]

\tikzstyle{leaf}=[my-box, minimum height=2em,
    fill=gray!5, text=black, align=left, font=\normalsize,
    inner xsep=3pt,
    inner ysep=6pt, 
    line width=0.5pt,
]

\definecolor{c1}{RGB}{102,178,255} 
\definecolor{c2}{RGB}{255,153,153} 
\definecolor{c3}{RGB}{255,204,102} 
\definecolor{c4}{RGB}{153,221,153} 
\definecolor{c5}{RGB}{204,179,255} 
\definecolor{c7}{RGB}{153,221,214} 
\definecolor{c8}{RGB}{221,160,221} 
\definecolor{c9}{RGB}{255,179,207} 

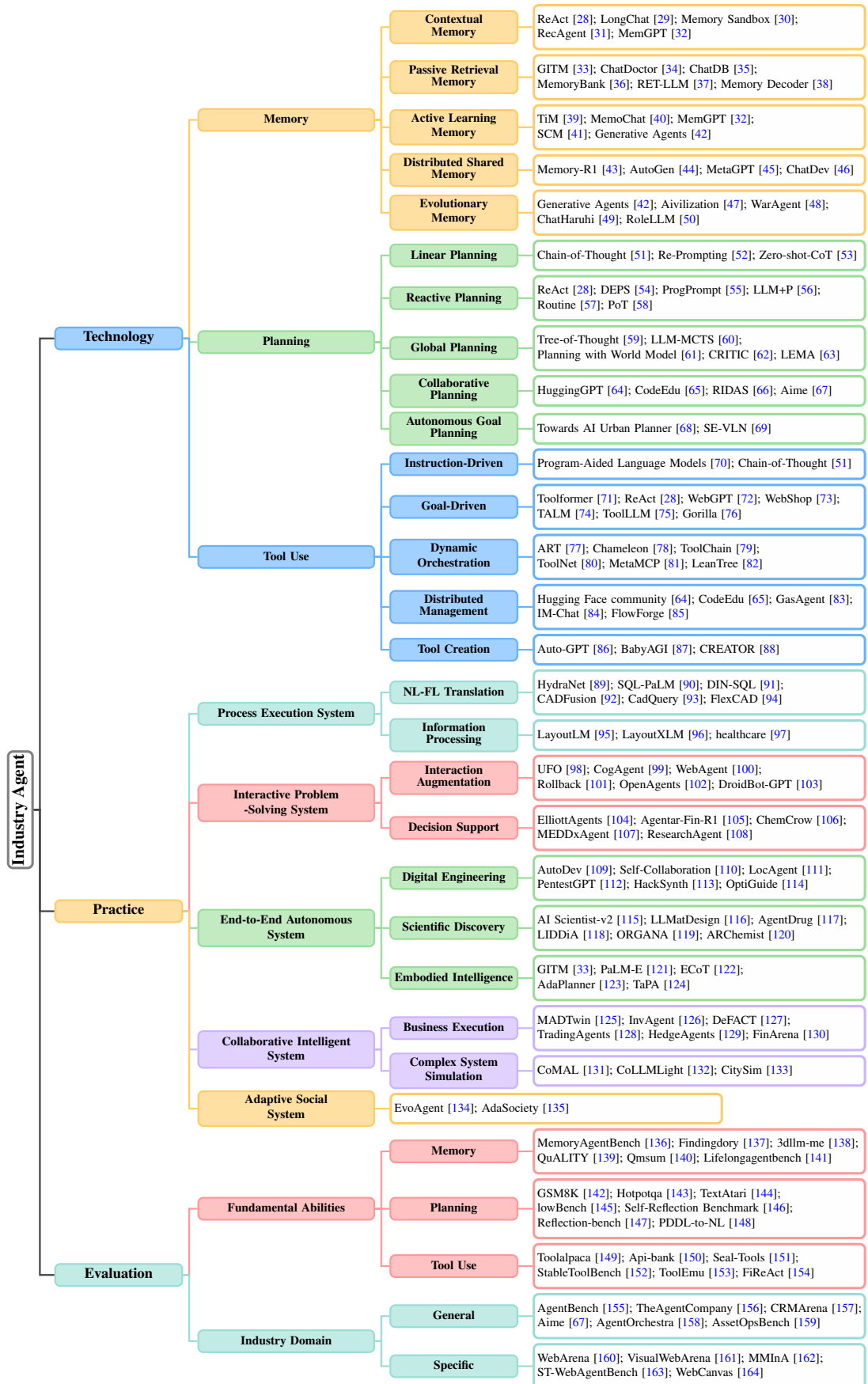
\begin{figure*}[!tp]
    \centering
    \resizebox{.8\textwidth}{!}{
        \begin{forest}
            forked edges,
            for tree={
                grow=east,
                reversed=true,
                anchor=base west,
                parent anchor=east,
                child anchor=west,
                base=center,
                font=\large,
                rectangle,
                draw=gray,
                rounded corners,
                align=left,
                text centered,
                minimum width=4em,
                edge+={darkgray, line width=0.5mm},
                s sep=3pt,
                inner xsep=2pt,
                inner ysep=3pt,
                line width=0.8pt,
                ver/.style={rotate=90, child anchor=north, parent anchor=south, anchor=center},
            },
            where level=1{text width=10em,font=\normalsize,}{},
            where level=2{text width=14em,font=\normalsize,}{},
            where level=3{text width=10em,font=\normalsize,}{},
            where level=4{text width=38em,font=\normalsize,}{}, 
            where level=5{text width=10em,font=\normalsize,}{},
            [
                \Large \textbf{Industry Agent}, ver, line width=0.7mm
                [
                    \large \shortstack{\textbf{Technology}}, fill=c1!60, draw=c1, line width=0.5mm
                    [
                        \textbf{Memory}, fill=c3!60, draw=c3, line width=0.5mm, edge={c3}
                        [
                            \shortstack{\textbf{Contextual} \\ \textbf{Memory}}, fill=c3!60, draw=c3, line width=0.5mm, edge={c3}
                            [
                                ReAct \cite{yao2023reactsynergizingreasoningacting};
                                LongChat \cite{LiHowLC};
                                Memory Sandbox \cite{huang2023memorysandboxtransparentinteractive};\\
                                RecAgent \cite{wang2024userbehaviorsimulationlarge};
                                MemGPT \cite{packer2024memgptllmsoperatingsystems}
                            ,leaf, text width=26.5em, draw=c3, line width=0.7mm, edge={c3} %
                            ]
                        ]
                        [
                            \shortstack{\textbf{Passive Retrieval} \\ \textbf{Memory}}, fill=c3!60, draw=c3, line width=0.5mm, edge={c3}
                            [
                                 GITM \cite{zhu2023ghostminecraftgenerallycapable};
                                 ChatDoctor \cite{li2023chatdoctormedicalchatmodel};
                                 ChatDB \cite{hu2023chatdbaugmentingllmsdatabases};\\
                                 MemoryBank \cite{zhong2023memorybankenhancinglargelanguage};
                                 RET-LLM \cite{modarressi2024retllmgeneralreadwritememory};
                                 Memory Decoder \cite{cao2025memorydecoderpretrainedplugandplay}
                            ,leaf, text width=26.5em, draw=c3, line width=0.7mm, edge={c3} %
                            ]
                        ]
                        [
                            \shortstack{\textbf{Active Learning} \\ \textbf{Memory}}, fill=c3!60, draw=c3, line width=0.5mm, edge={c3}
                            [
                                TiM \cite{liu2023thinkinmemoryrecallingpostthinkingenable};
                                MemoChat \cite{lu2023memochattuningllmsuse};
                                MemGPT \cite{packer2024memgptllmsoperatingsystems};\\
                                SCM \cite{wang2025scmenhancinglargelanguage};
                                Generative Agents \cite{park2023generativeagentsinteractivesimulacra}
                            ,leaf, text width=26.5em, draw=c3, line width=0.7mm, edge={c3} %
                            ]
                        ]
                        [
                            \shortstack{\textbf{Distributed Shared} \\ \textbf{Memory}}, fill=c3!60, draw=c3, line width=0.5mm, edge={c3}
                            [
                                Memory-R1 \cite{yan2025memoryr1enhancinglargelanguage};
                                AutoGen \cite{wu2023autogenenablingnextgenllm};
                                MetaGPT \cite{hong2024metagptmetaprogrammingmultiagent};
                                ChatDev \cite{qian2024chatdevcommunicativeagentssoftware}
                            ,leaf, text width=26.5em, draw=c3, line width=0.7mm, edge={c3} %
                            ]
                        ]
                        [
                            \shortstack{\textbf{Evolutionary} \\ \textbf{Memory}}, fill=c3!60, draw=c3, line width=0.5mm, edge={c3}
                            [
                            Generative Agents \cite{park2023generativeagentsinteractivesimulacra};
                            Aivilization \cite{al2024projectsidmanyagentsimulations};
                            WarAgent \cite{hua2024warpeacewaragentlarge};\\
                            ChatHaruhi \cite{li2023chatharuhirevivinganimecharacter};
                            RoleLLM \cite{wang2024rolellmbenchmarkingelicitingenhancing}
                            ,leaf, text width=26.5em, draw=c3, line width=0.7mm, edge={c3} %
                            ]
                        ]
                    ]
                    [
                        \shortstack{\textbf{Planning}}, 
                        fill=c4!60, draw=c4, line width=0.5mm, edge={c4}
                        [
                            \shortstack{\textbf{Linear Planning}}, fill=c4!60, draw=c4, line width=0.5mm, edge={c4}
                            [
                                Chain-of-Thought \cite{wei2023chainofthoughtpromptingelicitsreasoning};
                                Re-Prompting \cite{xu2024repromptingautomatedchainofthoughtprompt};
                                Zero-shot-CoT \cite{kojima2023largelanguagemodelszeroshot}
                            ,leaf, text width=26.5em, draw=c4, line width=0.7mm, edge={c4} %
                            ]
                        ]
                        [
                            \shortstack{\textbf{Reactive Planning}}, fill=c4!60, draw=c4, line width=0.5mm, edge={c4}
                            [
                                ReAct \cite{yao2023reactsynergizingreasoningacting};
                                DEPS \cite{wang2024describeexplainplanselect};
                                ProgPrompt \cite{singh2022progpromptgeneratingsituatedrobot};
                                LLM+P \cite{liu2023llmpempoweringlargelanguage};\\
                                Routine \cite{zeng2025routinestructuralplanningframework};
                                PoT \cite{chen2023programthoughtspromptingdisentangling}
                            ,leaf, text width=26.5em, draw=c4, line width=0.7mm, edge={c4} %
                            ]
                        ]
                        [
                            \shortstack{\textbf{Global Planning}}, fill=c4!60, draw=c4, line width=0.5mm, edge={c4}
                            [
                                Tree-of-Thought \cite{yao2023treethoughtsdeliberateproblem};
                                LLM-MCTS \cite{zheng2025montecarlotreesearch};\\
                                Planning with World Model \cite{hao2023reasoninglanguagemodelplanning};
                                CRITIC \cite{gou2024criticlargelanguagemodels};
                                LEMA \cite{an2024learningmistakesmakesllm}
                            ,leaf, text width=26.5em, draw=c4, line width=0.7mm, edge={c4} %
                            ]
                        ]
                        [
                            \shortstack{\textbf{Collaborative} \\ \textbf{Planning}}, fill=c4!60, draw=c4, line width=0.5mm, edge={c4}
                            [
                                HuggingGPT \cite{shen2023hugginggptsolvingaitasks};
                                CodeEdu \cite{zhao2025codeedumultiagentcollaborativeplatform};
                                RIDAS \cite{ding2025ridasmultiagentframeworkairan};
                                Aime \cite{shi2025aimefullyautonomousmultiagentframework}
                            ,leaf, text width=26.5em, draw=c4, line width=0.7mm, edge={c4} %
                            ]
                        ]
                        [
                            \shortstack{\textbf{Autonomous Goal} \\ \textbf{Planning}}, fill=c4!60, draw=c4, line width=0.5mm, edge={c4}
                            [
                             Towards AI Urban Planner \cite{fu2025urbanplaningaiagent};
                             SE-VLN \cite{dong2025sevlnselfevolvingvisionlanguagenavigation}
                            ,leaf, text width=26.5em, draw=c4, line width=0.7mm, edge={c4} %
                            ]
                        ]
                    ]
                    [
                        \shortstack{\textbf{Tool Use}}, fill=c1!60, draw=c1, line width=0.5mm, edge={c1}
                        [
                            \shortstack{\textbf{Instruction-Driven}}, fill=c1!60, draw=c1, line width=0.5mm, edge={c1}
                            [
                                Program-Aided Language Models \cite{gao2023palprogramaidedlanguagemodels};
                                Chain-of-Thought \cite{wei2023chainofthoughtpromptingelicitsreasoning}
                            ,leaf, text width=26.5em, draw=c1, line width=0.7mm, edge={c1} %
                            ]
                        ]
                        [
                            \shortstack{\textbf{Goal-Driven}}, fill=c1!60, draw=c1, line width=0.5mm, edge={c1}
                            [
                                Toolformer \cite{schick2023toolformerlanguagemodelsteach};
                                ReAct \cite{yao2023reactsynergizingreasoningacting};
                                WebGPT \cite{nakano2022webgptbrowserassistedquestionansweringhuman};
                                WebShop \cite{yao2023webshopscalablerealworldweb};\\
                                TALM \cite{parisi2022talmtoolaugmentedlanguage};
                                ToolLLM \cite{qin2023toolllmfacilitatinglargelanguage};
                                Gorilla \cite{patil2023gorillalargelanguagemodel}
                            ,leaf, text width=26.5em, draw=c1, line width=0.7mm, edge={c1} %
                            ]
                        ]
                        [
                            \shortstack{\textbf{Dynamic} \\ \textbf{Orchestration}}, fill=c1!60, draw=c1, line width=0.5mm, edge={c1}
                            [
                                ART \cite{paranjape2023artautomaticmultistepreasoning};
                                Chameleon \cite{chameleonteam2025chameleonmixedmodalearlyfusionfoundation};
                                ToolChain \cite{zhuang2023toolchainefficientactionspace};\\
                                ToolNet \cite{liu2024toolnetconnectinglargelanguage};
                                MetaMCP \cite{hou2025modelcontextprotocolmcp};
                                LeanTree \cite{kripner2025leantreeacceleratingwhiteboxproof}
                            ,leaf, text width=26.5em, draw=c1, line width=0.7mm, edge={c1} %
                            ]
                        ]
                        [
                            \shortstack{\textbf{Distributed} \\ \textbf{Management}}, fill=c1!60, draw=c1, line width=0.5mm, edge={c1}
                            [
                                Hugging Face community \cite{shen2023hugginggptsolvingaitasks};
                                CodeEdu \cite{zhao2025codeedumultiagentcollaborativeplatform};
                                GasAgent \cite{zheng2025gasagentmultiagentframeworkautomated};\\
                                IM-Chat \cite{lee2025imchatmultiagentllmbasedframework};
                                FlowForge \cite{hao2025flowforgeguidingcreationmultiagent}
                            ,leaf, text width=26.5em, draw=c1, line width=0.7mm, edge={c1} %
                            ]
                        ]
                        [
                            \shortstack{\textbf{Tool Creation}}, fill=c1!60, draw=c1, line width=0.5mm, edge={c1}
                            [
                            Auto-GPT \cite{yang2023autogptonlinedecisionmaking};
                            BabyAGI \cite{talebirad2023multiagentcollaborationharnessingpower};
                            CREATOR \cite{qian2024creatortoolcreationdisentangling}
                            ,leaf, text width=26.5em, draw=c1, line width=0.7mm, edge={c1} %
                            ]
                        ]
                    ]
                ]
                [   
                    \large \shortstack{\textbf{Practice}}, fill=c3!60, draw=c3, line width=0.5mm
                    [
                        \textbf{Process Execution System}, align=center, fill=c7!60, draw=c7, line width=0.5mm, edge={c7}
                        [
                            \shortstack{\textbf{NL-FL Translation}}, fill=c7!60, draw=c7, line width=0.5mm, edge={c7}
                            [
                                HydraNet~\cite{lyu2020hybridrankingnetworktexttosql};
                                SQL-PaLM~\cite{sun2024sqlpalmimprovedlargelanguage};
                                DIN-SQL~\cite{pourreza2023dinsqldecomposedincontextlearning};\\
                                CADFusion~\cite{wang2025texttocadgenerationinfusingvisual};
                                CadQuery~\cite{xie2025texttocadquerynewparadigmcad};
                                FlexCAD~\cite{zhang2025flexcadunifiedversatilecontrollable}
                            ,leaf, text width=26.5em, draw=c7, line width=0.7mm, edge={c7} %
                            ]
                        ]
                        [
                            \shortstack{\textbf{Information} \\ \textbf{Processing}}, fill=c7!60, draw=c7, line width=0.5mm, edge={c7}
                            [
                                LayoutLM~\cite{xu2020layoutlmpretrainingtextlayout};
                                LayoutXLM~\cite{xu2021layoutxlmmultimodalpretrainingmultilingual};
                                healthcare~\cite{kaiyrbekov2025automatedsurveycollectionllmbased}
                            ,leaf, text width=26.5em, draw=c7, line width=0.7mm, edge={c7} %
                            ]
                        ]
                    ]
                    [
                        \textbf{Interactive Problem} \\ \textbf{-Solving System}, align=center, fill=c2!60, draw=c2, line width=0.5mm, edge={c2}
                        [
                            \shortstack{\textbf{Interaction} \\ \textbf{Augmentation}}, fill=c2!60, draw=c2, line width=0.5mm, edge={c2}
                            [
                                UFO~\cite{zhang2024ufouifocusedagentwindows};
                                CogAgent~\cite{hong2024cogagentvisuallanguagemodel};
                                WebAgent~\cite{gur2024realworldwebagentplanninglong};\\
                                Rollback~\cite{zhang2025enhancingwebagentsexplicit};
                                OpenAgents~\cite{xie2023openagentsopenplatformlanguage};
                                DroidBot-GPT~\cite{wen2024droidbotgptgptpowereduiautomation}
                            ,leaf, text width=26.5em, draw=c2, line width=0.7mm, edge={c2} %
                            ]
                        ]
                        [
                            \shortstack{\textbf{Decision Support}}, fill=c2!60, draw=c2, line width=0.5mm, edge={c2}
                            [
                                ElliottAgents~\cite{chudziak2025elliottagentsnaturallanguagedrivenmultiagent};
                                Agentar-Fin-R1~\cite{zheng2025agentarfinr1enhancingfinancialintelligence};
                                ChemCrow~\cite{bran2023chemcrowaugmentinglargelanguagemodels};\\
                                MEDDxAgent~\cite{rose2025meddxagentunifiedmodularagent};
                                ResearchAgent~\cite{baek2025researchagentiterativeresearchidea}
                            ,leaf, text width=26.5em, draw=c2, line width=0.7mm, edge={c2} %
                            ]
                        ]
                    ]
                    [
                        \textbf{End-to-End Autonomous} \\ \textbf{System}, align=center, fill=c4!60, draw=c4, line width=0.5mm, edge={c4}
                        [
                            \shortstack{\textbf{Digital Engineering}}, fill=c4!60, draw=c4, line width=0.5mm, edge={c4}
                            [
                                AutoDev~\cite{tufano2024autodevautomatedaidrivendevelopment};
                                Self-Collaboration~\cite{dong2024selfcollaborationcodegenerationchatgpt};
                                LocAgent~\cite{chen2025locagentgraphguidedllmagents};\\
                                PentestGPT~\cite{deng2024pentestgptllmempoweredautomaticpenetration};
                                HackSynth~\cite{muzsai2024hacksynthllmagentevaluation};
                                OptiGuide~\cite{li2023largelanguagemodelssupply}
                            ,leaf, text width=26.5em, draw=c4, line width=0.7mm, edge={c4} %
                            ]
                        ]
                        [
                            \shortstack{\textbf{Scientific Discovery}}, fill=c4!60, draw=c4, line width=0.5mm, edge={c4}
                            [
                                AI Scientist-v2~\cite{yamada2025aiscientistv2workshoplevelautomated};
                                LLMatDesign~\cite{jia2024llmatdesignautonomousmaterialsdiscovery};
                                AgentDrug~\cite{le2025agentdrugutilizinglargelanguage};\\
                                LIDDiA~\cite{averly2025liddialanguagebasedintelligentdrug};
                                ORGANA~\cite{darvish2025organaroboticassistantautomated};
                                ARChemist~\cite{fakhruldeen2022archemistautonomousroboticchemistry}
                            ,leaf, text width=26.5em, draw=c4, line width=0.7mm, edge={c4} %
                            ]
                        ]
                        [
                            \shortstack{\textbf{Embodied Intelligence}}, fill=c4!60, draw=c4, line width=0.5mm, edge={c4}
                            [
                                GITM~\cite{zhu2023ghostminecraftgenerallycapable};
                                PaLM-E~\cite{driess2023palmeembodiedmultimodallanguage};
                                ECoT~\cite{zawalski2025roboticcontrolembodiedchainofthought};\\
                                AdaPlanner~\cite{sun2023adaplanneradaptiveplanningfeedback};
                                TaPA~\cite{wu2023embodiedtaskplanninglarge}
                            ,leaf, text width=26.5em, draw=c4, line width=0.7mm, edge={c4} %
                            ]
                        ]
                    ]
                    [
                        \textbf{Collaborative Intelligent} \\ \textbf{System}, align=center, fill=c5!60, draw=c5, line width=0.5mm, edge={c5}
                        [
                            \shortstack{\textbf{Business Execution}}, fill=c5!60, draw=c5, line width=0.5mm, edge={c5}
                            [
                                MADTwin~\cite{marah2024madtwin};
                                InvAgent~\cite{quan2025invagentlargelanguagemodel};
                                DeFACT~\cite{yang2024generative};\\
                                TradingAgents~\cite{xiao2025tradingagentsmultiagentsllmfinancial};
                                HedgeAgents~\cite{li2025hedgeagentsbalancedawaremultiagentfinancial};
                                FinArena~\cite{xu2025finarenahumanagentcollaborationframework}
                            ,leaf, text width=26.5em, draw=c5, line width=0.7mm, edge={c5} %
                            ]
                        ]
                        [
                            \shortstack{\textbf{Complex System} \\ \textbf{Simulation}}, fill=c5!60, draw=c5, line width=0.5mm, edge={c5}
                            [
                                CoMAL~\cite{yao2025comalcollaborativemultiagentlarge};
                                CoLLMLight~\cite{yuan2025collmlightcooperativelargelanguage};
                                CitySim~\cite{bougie2025citysim}
                            ,leaf, text width=26.5em, draw=c5, line width=0.7mm, edge={c5} %
                            ]
                        ]
                    ]
                    [
                         \shortstack{\textbf{Adaptive Social} \\ \textbf{System}}
                         , fill=c3!60, draw=c3, line width=0.5mm, edge={c3}
                        [
                                EvoAgent~\cite{yuan2025evoagent};
                                AdaSociety~\cite{huang2024adasociety}
                            ,leaf, text width=26.5em, draw=c3, line width=0.7mm, edge={c3} %
                        ]
                    ]
                ] 
                [   
                    \large \shortstack{\textbf{Evaluation}}, fill=c7!60, draw=c7, line width=0.5mm
                    [
                        \shortstack{\textbf{Fundamental Abilities}}, 
                        align=center, fill=c2!60, draw=c2, line width=0.5mm, edge={c2}
                        [
                            \shortstack{\textbf{Memory}}, fill=c2!60, draw=c2, line width=0.5mm, edge={c2}
                            [
                                MemoryAgentBench \cite{hu2025evaluatingmemoryllmagents};
                                Findingdory \cite{yadav2025findingdorybenchmarkevaluatememory};
                                3dllm-me \cite{hu20253dllmmemlongtermspatialtemporalmemory}; \\
                                QuALITY  \cite{pang2022qualityquestionansweringlong};
                                Qmsum  \cite{zhong2021qmsumnewbenchmarkquerybased};
                                Lifelongagentbench \cite{zheng2025lifelongagentbenchevaluatingllmagents}
                            ,leaf, text width=26.5em, draw=c2, line width=0.7mm, edge={c2} %
                            ]
                        ]
                        [
                            \shortstack{\textbf{Planning}}, fill=c2!60, draw=c2, line width=0.5mm, edge={c2}
                            [
                                GSM8K  \cite{cobbe2021trainingverifierssolvemath};
                                Hotpotqa \cite{yang2018hotpotqadatasetdiverseexplainable};
                                TextAtari \cite{li2025textatari100kframesgame};\\
                                lowBench \cite{xiao2024flowbenchrevisitingbenchmarkingworkflowguided};
                                Self-Reflection Benchmark \cite{renze2024selfreflectionllmagentseffects};\\
                                Reflection-bench \cite{li2025reflectionbenchevaluatingepistemicagency};
                                PDDL-to-NL \cite{stein2025automatinggenerationpromptsllmbased}
                            ,leaf, text width=26.5em, draw=c2, line width=0.7mm, edge={c2} %
                            ]
                        ]
                        [
                            \shortstack{\textbf{Tool Use}}, fill=c2!60, draw=c2, line width=0.5mm, edge={c2}
                            [
                                Toolalpaca \cite{tang2023toolalpacageneralizedtoollearning};
                                Api-bank \cite{li2023apibankcomprehensivebenchmarktoolaugmented};
                                Seal-Tools \cite{wu2024sealtoolsselfinstructtoollearning};\\
                                StableToolBench \cite{guo2025stabletoolbenchstablelargescalebenchmarking};
                                ToolEmu  \cite{ruan2024identifyingriskslmagents};
                                FiReAct  \cite{müller2025semanticcontexttoolorchestration}
                            ,leaf, text width=26.5em, draw=c2, line width=0.7mm, edge={c2} %
                            ]
                        ]
                    ]
                    [
                        \shortstack{\textbf{Industry Domain}}, 
                        align=center, fill=c7!60, draw=c7, line width=0.5mm, edge={c7}
                        [
                            \shortstack{\textbf{General}}, fill=c7!60, draw=c7, line width=0.5mm, edge={c7}
                            [
                                AgentBench~\cite{liu2023agentbenchevaluatingllmsagents};
                                TheAgentCompany~\cite{xu2025theagentcompanybenchmarkingllmagents};
                                CRMArena~\cite{huang2025crmarenaunderstandingcapacityllm};\\
                                Aime\cite{shi2025aimefullyautonomousmultiagentframework};
                                AgentOrchestra~\cite{zhang2025agentorchestrahierarchicalmultiagentframework};
                                AssetOpsBench~\cite{patel2025assetopsbenchbenchmarkingaiagents}
                            ,leaf, text width=26.5em, draw=c7, line width=0.7mm, edge={c7} %
                            ]
                        ]
                        [
                            \shortstack{\textbf{Specific}}, fill=c7!60, draw=c7, line width=0.5mm, edge={c7}
                            [
                                WebArena~\cite{zhou2024webarenarealisticwebenvironment};
                                VisualWebArena~\cite{he2024webvoyagerbuildingendtoendweb};
                                MMInA~\cite{tian2025mminabenchmarkingmultihopmultimodal};\\
                                ST-WebAgentBench\cite{levy2025stwebagentbenchbenchmarkevaluatingsafety};
                                WebCanvas~\cite{pan2024webcanvasbenchmarkingwebagents}
                            ,leaf, text width=26.5em, draw=c7, line width=0.7mm, edge={c7} %
                            ]
                        ]
                    ]
                ]
            ]
        \end{forest}}
    \caption{Taxonomy of industry agents.}
    \label{fig:taxonomy}
\end{figure*}

\section{Technical Foundations of Industry Agents}

In recent years, agents built upon LLMs have made significant advancements. Their increasingly sophisticated capabilities in handling complex tasks are steering artificial intelligence research and applications toward higher levels of cognitive intelligence. Early agent research was often limited to specific tasks. In contrast, emerging LLMs, with their robust general language understanding, reasoning, and interaction abilities, have greatly facilitated the emergence of general-purpose agents capable of handling open-domain complex tasks.

Currently, a comprehensive general-purpose agent framework typically relies on three core technical pillars: Memory, Planning, and Tool Use. Memory refers to the ability to encode, store, and retrieve information; Planning involves goal decomposition and the formulation and optimization of action sequences; Tool Use pertains to the ability to invoke external APIs or programs to extend one's capabilities. These three core modules are interwoven and collaborate, forming the foundation for agents to perceive their environment, develop cognition, and take action. This enables agents to evolve from simple instruction executors to autonomous entities capable of continuous interaction with their environment and achieving complex objectives.

However, as agent research increasingly covers real-world scenarios, cognitive bottlenecks in their core architectures have become more apparent. These challenges are deeply reflected in the limitations of the three core capabilities: Memory, Planning, and Tool Use. In the realm of Memory, limited and singular context windows make it difficult for agents to maintain long-term, coherent interaction histories, leading to issues like long-context forgetting. Additionally, how to filter, refine, and form structured, effective memories from vast, noisy, unstructured dynamic environmental information, avoiding information overload and cognitive biases, remains a significant technical bottleneck. In Planning, the high dynamism and uncertainty of the real world render simple planning methods based on static world assumptions ineffective. Agents must possess the robustness to dynamically adjust plans during execution, handle anomalies, and learn from failures, placing high demands on their ability to decompose long-term goals and reason effectively. Regarding Tool Use, as tool libraries grow large and complex, how to accurately select, combine, and invoke appropriate tools to solve problems, as well as how to handle tool execution failures or unexpected results, become critical factors limiting the upper bounds of agent capabilities. These practical technical challenges collectively form a gap between theoretical frameworks and real-world applications.

To systematically analyze how industry agents evolve from simple process automation tools to core systems capable of solving complex domain problems, this review proposes a five-level framework (L1–L5) oriented toward industry application capability maturity. This framework aims to reveal that the transitions at each capability level of industry agents are essentially driven by their evolution in the three core technologies: Memory, Planning, and Tool Use. For instance, the Process Execution System at the L1 level requires only transient memory and fixed linear planning, whereas the Adaptive Social System at the L5 level demands the ability to accumulate evolutionary group memory across generations and the capacity to autonomously generate goals in complex games. The following sections will delve into each of the three core technical modules, analyzing how their technological evolution supports the continuous upgrading of industry agent capabilities, thereby laying the groundwork for the development practices of industry agents.

\begin{figure}[!t]
  \centering
  \includegraphics[width=.9\linewidth]{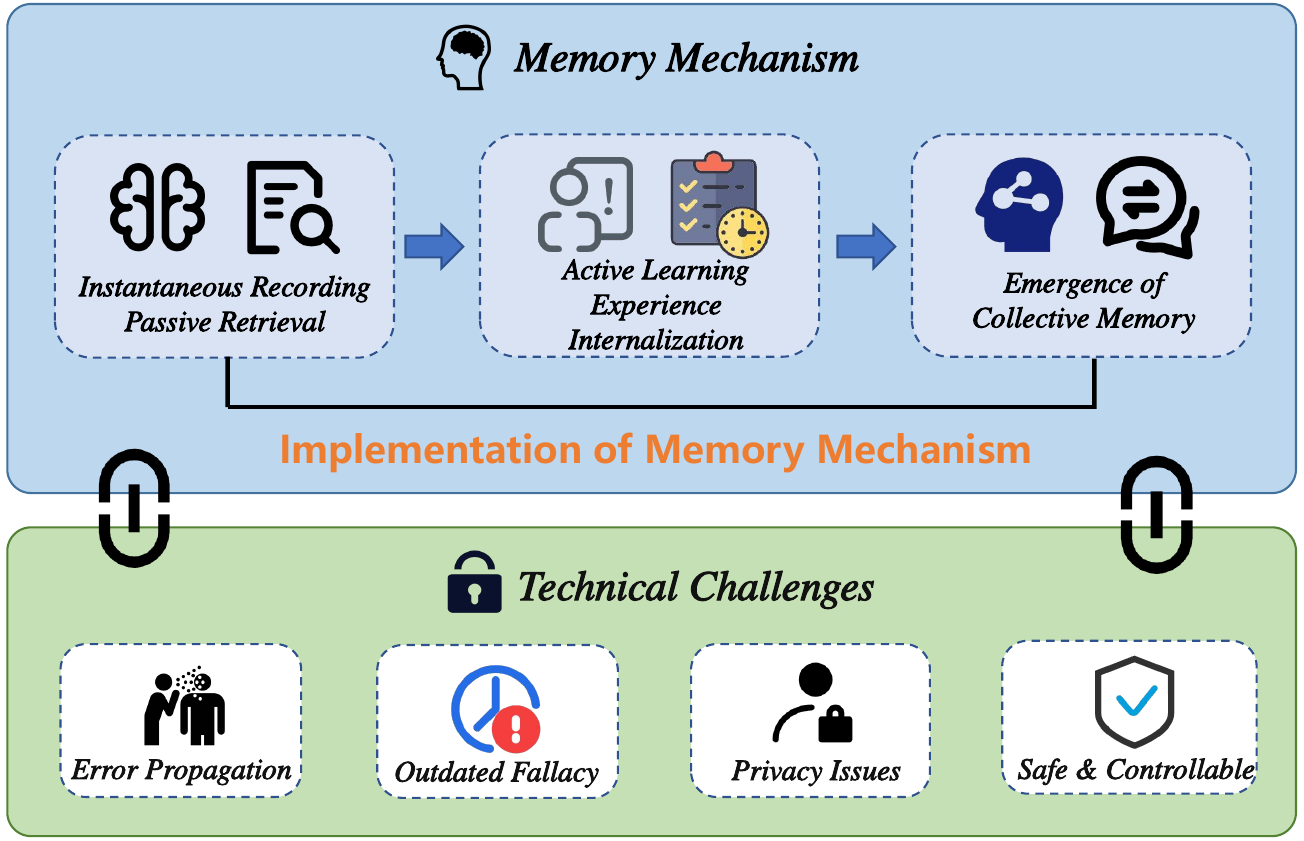}
  \caption{The evolution of memory mechanisms in industry agents.}
  \label{fig: memory}
\end{figure}

\subsection{Memory Mechanism}

Memory is a core component in building advanced artificial intelligence, particularly in LLM-based agents. It enables agents to encode, store, and retrieve historical information, allowing them to transcend the stateless limitations of traditional computational models and exhibit the ability to learn, adapt, and execute complex tasks coherently. For industry agents aimed at solving real-world problems, the complexity and maturity of their memory mechanisms determine the capability levels and application value they can achieve within specific domains. Previous reviews often categorize existing works based on technical implementations, such as the sources, forms, or operations of memory. While such classifications aid in understanding technical details, they do not fully reveal how the evolution of memory mechanisms directly drives the capability transitions of agents. This section analyzes how memory, as a core technology, evolves from supporting basic process execution to underpinning autonomous learning and even collective collaboration in complex systems.

\subsubsection{From Instantaneous Recording to Passive Retrieval}

In the early stages of industry applications, the core value of agents lies in processing explicit instructions and utilizing existing knowledge. Their memory mechanisms primarily focus on two basic functions: recording and querying. This phase marks the transition of agents from being stateless executors to assistants capable of consulting external notes.

At L1, memory is instantaneous context, essentially working memory. In this phase, agents function as process execution systems, with their memory capabilities mainly supported by the context window of LLMs. This memory is temporary and task-oriented, used solely to maintain information consistency within a single interaction, akin to human short-term working memory. The chain of thought in the ReAct framework exemplifies this instantaneous working memory, explicitly retaining the reasoning process within the context to guide subsequent actions \cite{yao2023reactsynergizingreasoningacting}. However, the fundamental limitation of this memory lies in its finiteness and volatility. To extend this limited memory capacity, works like LongChat fine-tune the base model to better handle and remember longer, complete interactions \cite{LiHowLC}. Yet, longer contexts may introduce interference. To address this, Memory Sandbox designs an interactive memory management interface, allowing manual removal of irrelevant information before feeding memory into prompts, reflecting an initial attempt at controlling the quality of instantaneous memory \cite{huang2023memorysandboxtransparentinteractive}. Some systems have designed more structured short-term memories, such as the short-term memory cache in RecAgent \cite{wang2024userbehaviorsimulationlarge} for recommendation scenarios and the flash memory or working context in MemGPT \cite{packer2024memgptllmsoperatingsystems} for holding recent interaction histories. These can be seen as optimizations of instantaneous memory but do not alter its nature of being forgotten after task completion.

At L2, memory evolves to passive retrieval, marking the emergence of long-term memory. When agents function as interactive problem-solving systems, relying solely on context memory is insufficient to handle queries requiring domain knowledge. Therefore, a key evolution of memory mechanisms is the connection to external knowledge bases, achieving a transition from stateless to knowledgeable. The core of memory at this stage is retrieval-augmented generation (RAG), enabling agents to passively retrieve information from external sources to enhance their responses \cite{lewis2021retrievalaugmentedgenerationknowledgeintensivenlp}. Works like Toolformer \cite{schick2023toolformerlanguagemodelsteach} and ToolLLM \cite{qin2023toolllmfacilitatinglargelanguage} teach models how to use tools and thousands of real APIs, laying the foundation for acquiring external knowledge. In general scenarios, ReAct demonstrates how to call the Wikipedia API \cite{yao2023reactsynergizingreasoningacting} . In industry applications, this retrieval becomes more targeted: in software engineering, CodeAgent \cite{zhang2024codeagentenhancingcodegeneration} designs web search strategies to solve code dependency issues; in gaming, GITM \cite{zhu2023ghostminecraftgenerallycapable} draws knowledge from the online Minecraft Wiki; in scientific computing, ToRA \cite{gou2024toratoolintegratedreasoningagent} enhances agents' ability to use programmatic tools, while ChemCrow \cite{bran2023chemcrowaugmentinglargelanguagemodels} equips LLMs with chemical tools; in professional Q\&A, ChatDoctor \cite{li2023chatdoctormedicalchatmodel} fine-tunes retrieval models to obtain knowledge from Wikipedia and medical databases. These external knowledge bases constitute the nascent form of stable and reliable long-term memory for agents. Technologies for efficient retrieval are also becoming more diverse. For example, ChatDB \cite{hu2023chatdbaugmentingllmsdatabases} generates SQL queries for precise retrieval from structured databases, MemoryBank \cite{zhong2023memorybankenhancinglargelanguage} uses dual-tower dense retrieval models, and RET-LLM \cite{modarressi2024retllmgeneralreadwritememory} employs locality-sensitive hashing for fast reading. Memory Decoder \cite{cao2025memorydecoderpretrainedplugandplay} is a plug-and-play pre-trained memory component that mimics the behavior of external non-parametric retrievers to achieve efficient domain adaptation, enhancing performance in specialized fields without modifying the original model parameters. At this stage, although agents possess long-term memory, the memory is static and external. The agents themselves have not learned the knowledge; they are merely more efficient queryers, with their capability boundaries limited by the quality of external knowledge bases and the precision of retrieval algorithms.

\subsubsection{Active Learning and Experience Internalization}

At L3, agents evolve into end-to-end systems capable of completing complex tasks in a closed-loop manner. This progression signifies a fundamental shift in their memory mechanisms. Memory transitions from passive information storage and retrieval to an active, dynamic system that facilitates learning and experience internalization. Central to this advancement is the agent's acquisition of metacognitive abilities, enabling self-reflection and the extraction of actionable insights from experiences.

A defining characteristic of this stage is the agent's capacity for active learning from its interaction history. These experiences stem from various sources. Some arise from observations during individual task executions, such as in Generative Agents \cite{park2023generativeagentsinteractivesimulacra}, where agents document all their actions in a simulated world, or in Voyager \cite{wang2023voyageropenendedembodiedagent}, which records reusable code executed successfully in Minecraft. More valuable experiences emerge from analyzing the successes and failures across multiple task attempts. The Reflexion framework introduces verbal reinforcement learning, allowing agents to reflect on their actions and store outcomes in memory \cite{shinn2023reflexionlanguageagentsverbal}. Retroformer enhances this by fine-tuning reflection models for more effective cross-experiment information extraction \cite{yao2024retroformerretrospectivelargelanguage}. ExpeL \cite{zhao2024expelllmagentsexperiential} and Synapse \cite{zheng2024synapsetrajectoryasexemplarpromptingmemory} utilize successful task trajectories as exemplars, retrieving similar past cases to guide new tasks. In this phase, agents transcend being mere information consumers; they become producers and distillers of experience. Through reflection, they transform disparate, first-order interaction records into structured, higher-order action guidelines, laying the cognitive foundation for autonomous improvement and long-term task planning.

Moreover, L3 agents not only retrieve experiences but also internalize them, converting external knowledge into internal memory. Traditional internalization methods involve fine-tuning. For instance, Character-LLM fine-tunes on role-specific data like scripts to embed character traits into model parameters \cite{shao2023characterllmtrainableagentroleplaying}. In specialized domains, Huatuo \cite{wang2023huatuotuningllamamodel}, DoctorGLM \cite{xiong2023doctorglmfinetuningchinesedoctor}, and Radiology-GPT \cite{liu2024radiologygptlargelanguagemodel} fine-tune on Chinese medical knowledge, medical data, and radiology datasets, respectively, endowing agents with professional biomedical knowledge. InvestLM fine-tunes for financial investment capabilities \cite{yang2023investlmlargelanguagemodel}. Beyond fine-tuning, more refined memory editing techniques are emerging, allowing modification of specific knowledge in model parameters without retraining. MEND \cite{li2024mendmetademonstrationdistillation} and KnowledgeEditor \cite{decao2021editingfactualknowledgelanguage} train lightweight editing networks to predict parameter updates for rapid factual knowledge modifications. MAC \cite{tack2024onlineadaptationlanguagemodels} employs meta-learning for online parameterized memory adaptation, while PersonalityEdit \cite{mao2024editingpersonalitylargelanguage} enables precise editing of agent personality traits based on psychological theories like the Big Five personality model. Traditional fine-tuning and knowledge editing represent a profound shift from knowledge storage to ability cultivation, serving as technological prerequisites for end-to-end autonomous systems. However, they are often uncontrollable and less interpretable.

In the realm of memory, a more general method of internalization is non-parametric memory management, encompassing: 1) Memory Writing: Efficiently storing raw information into memory. TiM \cite{liu2023thinkinmemoryrecallingpostthinkingenable} extracts information into entity relationships stored in structured databases; MemoChat \cite{lu2023memochattuningllmsuse} summarizes dialogue segments into themes as indexed keys; MemGPT \cite{packer2024memgptllmsoperatingsystems} implements self-guided memory updates; SCM \cite{wang2025scmenhancinglargelanguage} designs memory controllers to determine when to perform write operations. 2) Memory Management and Refinement: Extracting value from vast memories and preventing degradation. Generative Agents \cite{park2023generativeagentsinteractivesimulacra} generate higher-level abstract thinking through reflection processes; MemoryBank \cite{zhong2023memorybankenhancinglargelanguage} distills daily conversations into high-level daily summaries; GITM \cite{zhu2023ghostminecraftgenerallycapable} summarizes key actions from multiple plans; Voyager \cite{wang2023voyageropenendedembodiedagent} optimizes and refines its skill library based on environmental feedback (e.g., code execution success). 3) Memory Reading: Retrieving the most relevant memories based on current tasks. ChatDB \cite{hu2023chatdbaugmentingllmsdatabases} generates SQL queries for retrieval; ExpeL \cite{zhao2024expelllmagentsexperiential} uses the Faiss vector library to retrieve the top-K most similar successful trajectories; MPC \cite{lee2023promptedllmschatbotmodules} provides chain-of-thought examples to guide models in ignoring irrelevant memories. To integrate these processes, Mem0 adopts a scalable memory center architecture, dynamically extracting, integrating, and retrieving key information from dialogues, significantly enhancing long-dialogue consistency and reducing computational overhead \cite{chhikara2025mem0buildingproductionreadyai}. The upgraded Mem0 represents memory as an entity-relationship graph, capturing complex temporal and multi-hop reasoning logic, particularly suited for cross-session and cross-timepoint conversations.  draws inspiration from the Zettelkasten method, employing atomic note structures, multi-dimensional semantic representations, and a combination of vector retrieval and LLM analysis to achieve automatic linking and self-evolution of memory.

\subsubsection{Emergence of Collective Memory}

At L4 and above, industry applications expand from single-agent systems to complex multi-agent collaborations, leading to the emergence of collective memory.

Level 4 memory is distributed and shared. When multiple agents collaborate to achieve a large-scale goal, they must rely on a shared cognitive space, forming their collective memory. Memory-R1 enables large models to actively manage and utilize external memory through alternating operations between two agents \cite{yan2025memoryr1enhancinglargelanguage}. Additionally, multi-agent frameworks such as AutoGen \cite{wu2023autogenenablingnextgenllm}, ChatDev \cite{qian2024chatdevcommunicativeagentssoftware}, and MetaGPT \cite{hong2024metagptmetaprogrammingmultiagent} provide implementation examples. In these systems, all agent roles share a unified context, including requirement documents, codebases, and API specifications. For instance, in the ChatDev simulation of software development, each agent stores past dialogues with other roles \cite{qian2024chatdevcommunicativeagentssoftware}. In MetaGPT, agents can retrieve historical records from memory to address errors \cite{hong2024metagptmetaprogrammingmultiagent}. In broader collaborative scenarios, such as the S³ social network simulation, each agent's memory pool contains diverse user messages to define its identity \cite{gao2025s3socialnetworksimulationlarge}. In the job simulation MetaAgents, memory continuously enriches through dialogue and reflection \cite{li2025metaagentslargelanguagemodel}. In the code repair scenario RTLFixer, an externally shared database stores compiler errors and expert repair instructions \cite{tsai2024rtlfixerautomaticallyfixingrtl}. Shared memory serves as the foundation for efficient collaboration, ensuring all individuals communicate and cooperate based on consistent information, thereby avoiding information silos and cognitive biases. It is a prerequisite for coordinating complex business processes. Despite various multi-agent communication protocols and topologies being proposed, this memory remains synchronous and task-oriented.

Level 5 memory envisions an evolutionary and cultural form. At this level, memory not only shares across agents but also accumulates, solidifies, and evolves over time, forming a culture akin to human societies. It records the group's successful strategies, lessons from failures, and shared values, which can be inherited by newly joined agents. In Generative Agents, information propagation within the agent society demonstrates this primitive form of memory \cite{park2023generativeagentsinteractivesimulacra}. Aivilization scales this concept to a larger community, constructing a highly realistic virtual society encompassing economics, industry, politics, and social interactions \cite{al2024projectsidmanyagentsimulations}. In specific simulations, such as WarAgent's war simulation, dialogues of participating countries are continuously recorded in memory, shaping their long-term behaviors \cite{hua2024warpeacewaragentlarge}. In role-playing applications like ChatHaruhi \cite{li2023chatharuhirevivinganimecharacter} and RoleLLM \cite{wang2024rolellmbenchmarkingelicitingenhancing}, by injecting role-specific knowledge and plot memories, agents exhibit consistent identities, reflecting micro-level cultural forms. Exploring L5 memory involves considering how to build an agent society capable of self-improvement and sustainable development. This requires memory mechanisms that not only record "what is" but also encapsulate "why it is" and "how it should be," thereby providing a foundation for long-term value alignment and goal evolution within agent communities.

\subsubsection{Real-World Challenges in Memory Management}

Efficient memory mechanisms do not equate to flawless performance. Empirical studies by \cite{xiong2025memorymanagementimpactsllm}. reveal that LLM agents exhibit a pronounced experience-following behavior, meaning they tend to replicate past experiences similar to current tasks. This characteristic introduces two primary risks: error propagation, where mistakes in early memories are amplified in subsequent decisions; and misaligned experience replay, where outdated or irrelevant memories negatively interfere with the current task. Their research emphasizes the importance of implementing refined memory addition and deletion strategies to maintain long-term agent robustness. Their research emphasizes the importance of implementing refined memory addition and deletion strategies to maintain long-term agent robustness. Concurrently, Wang et al. systematically identify privacy vulnerabilities within agent memory modules \cite{wang2025unveilingprivacyrisksllm}. They introduce the Memory EXTRaction Attack (MEXTRA), demonstrating that even in black-box settings, attackers can extract sensitive user interactions stored in memory through prompt engineering. Their findings underscore the necessity for secure and controllable memory systems, especially in high-risk, high-regulation sectors such as healthcare, finance, and law. These studies collectively highlight a central challenge in memory research: the need to develop memory systems that are not only effective in learning but also secure, controllable, and maintainable.

\begin{figure}[!t]
  \centering
  \includegraphics[width=.9\linewidth]{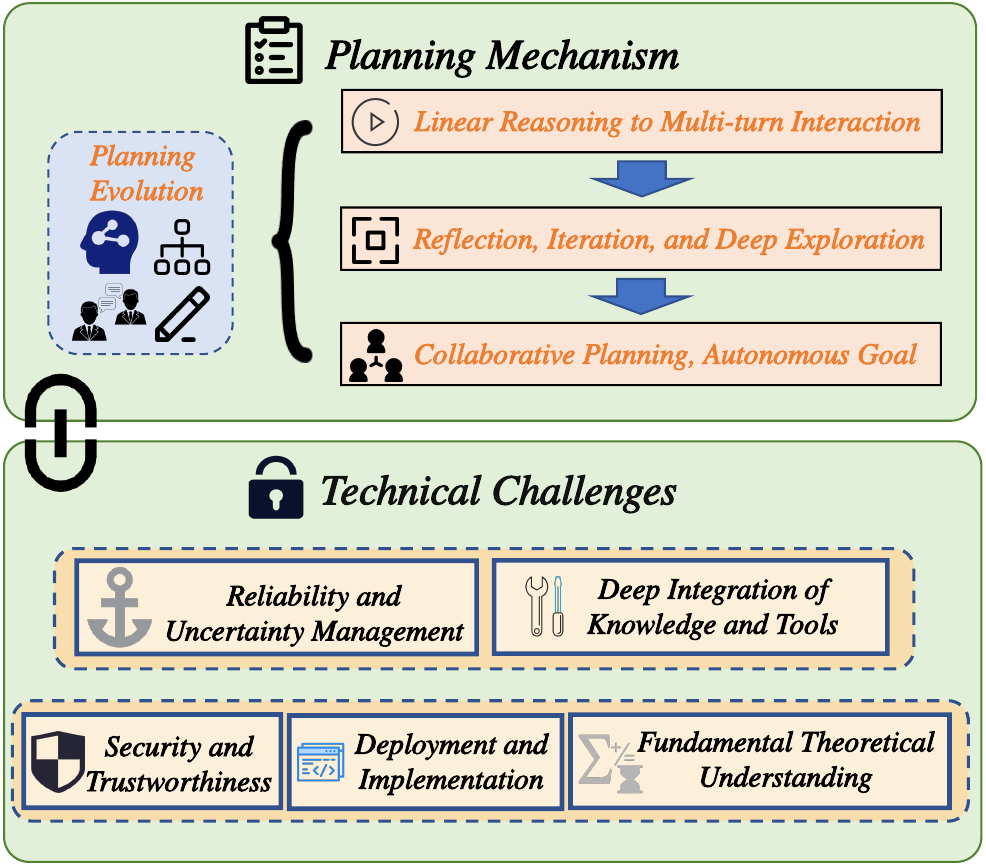}
  \caption{The evolution of planning capability in industry agents.}
  \label{fig: planning}
\end{figure}

\subsection{Planning Capability}

Planning is a core cognitive ability of an agent. It determines how an agent decomposes abstract goals into a series of executable actions to achieve its intentions in an environment. In the context of industry agents, planning capability is directly related to their autonomy, reliability, and the complexity of problem-solving. A robust planning module enables an agent not only to understand "what to do" but also to autonomously decide "how to do it," adjusting and optimizing in dynamic environments. This section systematically reviews how planning technologies have evolved from simple task decomposition to complex reflective, collaborative, and generative planning, driving industry agents to create core value at different maturity stages. Each leap in planning capability marks a foundational step toward higher autonomy and intelligence in agents.

\subsubsection{From Linear Reasoning to Multi-turn Interaction}

At the L1-L2 stages, agents primarily serve as human assistants or tools, with their planning capabilities focusing on accurately understanding user instructions and decomposing them into executable steps.
At the L1 level, planning is linear instruction decomposition, essentially open-loop planning. The core involves following a relatively fixed, pre-set path to complete tasks. A breakthrough in this stage is the Chain-of-Thought (CoT) prompting technique, which guides LLMs to generate intermediate reasoning steps, significantly enhancing their ability to handle complex problems \cite{wei2023chainofthoughtpromptingelicitsreasoning}. However, CoT is inherently a linear, one-time generation process, lacking interaction and correction with the environment. Re-Prompting improves this by utilizing precondition error information to re-prompt LLMs, generating executable plans that enhance the plans' executability and semantic correctness \cite{xu2024repromptingautomatedchainofthoughtprompt}. The emergence of Zero-shot-CoT further simplifies this process, requiring only a phrase like "Let's think step by step" to trigger the model's initial reasoning ability \cite{kojima2023largelanguagemodelszeroshot}. Plan-and-Solve Prompting structures this process by explicitly dividing the task into "plan first, then execute" steps, laying the foundation for more reliable execution \cite{wang2023planandsolvepromptingimprovingzeroshot}. Linear planning enables agents to handle multi-step logical problems but assumes a static environment and flawless initial planning, which often does not hold in the ever-changing real world, leading to reduced robustness.

At the L2 level, planning evolves into reactive planning, achieving closed-loop control. Agents are no longer merely passively decomposing tasks but can interact with the environment or tools and dynamically adjust subsequent steps based on feedback. 

The ReAct framework is a milestone at this stage, decoupling and interleaving reasoning and action, allowing agents to think during execution and act after thinking \cite{yao2023reactsynergizingreasoningacting}. This "think-act-observe" cycle forms the core of reactive planning. This mode is particularly crucial in scenarios requiring interaction with external tools. For example, Visual ChatGPT utilizes the ReAct mechanism, using an LLM as the brain to orchestrate a series of visual foundation models to complete image processing tasks \cite{yao2023reactsynergizingreasoningacting}. DEPS enhances the task planning capability of multi-task agents in open-world environments by combining descriptions of the planning execution process, self-explanations upon failure, and a trainable goal selector that estimates completion steps to parallel subgoals \cite{wang2024describeexplainplanselect}. To make the planning process more rigorous and predictable, researchers have explored converting natural language planning into more formal languages. PAL \cite{gao2023palprogramaidedlanguagemodels} and Program-of-Thought Prompting (PoT) \cite{chen2023programthoughtspromptingdisentangling} guide LLMs to express reasoning processes as executable code, utilizing the determinism of code interpreters to ensure result accuracy. ProgPrompt adopts a similar approach, transforming robotic task planning into function generation problems \cite{singh2022progpromptgeneratingsituatedrobot}. Furthermore, to meet the high reliability requirements of industrial applications, a series of works combine LLMs with classical symbolic planners. Frameworks like LLM+P \cite{liu2023llmpempoweringlargelanguage}, LLM+PDDL \cite{guan2023leveragingpretrainedlargelanguage}, and LLM+ASP \cite{yang2023couplinglargelanguagemodels} use LLMs to convert natural language problems into formal representations such as PDDL or ASP , then call external optimization planners to solve them, obtaining optimal and reliable plans. In enterprise environments, the Routine framework achieves stable multi-step tool invocation planning by providing clear structures and instructions \cite{zeng2025routinestructuralplanningframework}. In gaming scenarios, Voyager achieves continuous exploration, skill acquisition, and autonomous discovery in a human-free Minecraft environment through automatic curriculum planning \cite{wang2023voyageropenendedembodiedagent}. Reactive planning greatly enhances an agent's adaptability in dynamic environments but typically has a localized, short-sighted planning perspective. It excels at "adapting to changes" but struggles with "deep thinking," making it challenging to solve complex problems requiring long-term planning and trade-offs.

\subsubsection{Global Planning — Reflection, Iteration, and Deep Exploration}

As agents advance to the L3 level, becoming "end-to-end autonomous systems," their planning capabilities must address complex, dynamic, and uncertain environments. This necessitates planning processes with nonlinear abilities for deep exploration, self-correction, and continuous learning.

As agents advance to the L3 level, becoming end-to-end autonomous systems, their planning capabilities must address complex, dynamic, and uncertain environments. This necessitates planning processes with nonlinear abilities for deep exploration, self-correction, and continuous learning.
Initially, planning evolves from linear chains to tree-like or graph-like explorations. Compared to CoT \cite{wei2023chainofthoughtpromptingelicitsreasoning}, Tree-of-Thought (ToT) \cite{yao2023treethoughtsdeliberateproblem} and Graph-of-Thought (GoT) \cite{besta2024graphthoughtssolvingelaborate} significantly expand the planning exploration space. ToT organizes reasoning paths into a tree structure, allowing agents to explore, evaluate, and even backtrack among multiple potential solutions to select the globally optimal path. GoT further models the thought process as a graph, supporting more complex thought aggregation and transformation, thereby enhancing the ability to solve intricate problems.

To navigate vast search spaces efficiently, frameworks like LLM-MCTS \cite{zheng2025montecarlotreesearch} and Reasoning with Language Model is Planning with World Model (RAP) \cite{hao2023reasoninglanguagemodelplanning} innovatively utilize LLMs as heuristic functions in Monte Carlo Tree Search (MCTS) \cite{Swiechowski2023MCTSReview}, guiding the search process to balance exploration and exploitation. By introducing systematic search strategies, agents evolve from greedy decision-makers to ones capable of trade-offs and foresight, which is crucial for solving complex problems with multiple potential paths and pitfalls.
Self-reflection and correction become core mechanisms, transforming planning from a one-time process into an iterative optimization cycle. Agents are no longer one-off planners; they possess the ability to learn from experiences and failures.

The Reflexion framework builds upon ReAct by adding a self-reflection loop, enabling agents to analyze failure trajectories, generate textual reflections, and store them in memory to guide subsequent attempts \cite{shinn2023reflexionlanguageagentsverbal}. Self-Refine introduces an iterative optimization process without external training, where agents generate solutions, provide feedback on them, and refine the solutions in subsequent rounds \cite{madaan2023selfrefineiterativerefinementselffeedback}. The CRITIC framework employs external tools, such as knowledge bases and search engines, to verify and critique agent-generated actions, using external feedback for self-correction \cite{gou2024criticlargelanguagemodels}. The LEMA framework collects erroneous planning samples, utilizes more powerful models for corrections, and fine-tunes the original model with these corrected samples \cite{an2024learningmistakesmakesllm}.

In more specific applications, such as formal mathematical proofs, the Delta Prover framework iteratively constructs proofs through interactions, reflections, and reasoning between LLMs and the proof environment Lean 4 \cite{zhou2025solvingformalmathproblems}. In robotics, the Conditional Multi-Stage Failure Recovery framework designs multi-stage failure recovery strategies for embodied agents, enhancing their robustness in executing tasks in real-world environments \cite{farag2025conditionalmultistagefailurerecovery}.

Comprehensive and macro-level reflection mechanisms empower agents to draw lessons from errors. This endows planning with resilience, enabling agents to recover from mistakes and continuously improve, which is key to achieving end-to-end autonomy.

\subsubsection{Collaborative Planning and Autonomous Goal Setting}

At L4 and beyond, the scope of planning extends from individual agents to systems composed of multiple agents, broadening the concept of planning from task execution to collaborative strategies and social evolution.

At the L4 level, planning focuses on how multiple agents can develop and execute group plans through communication and negotiation to achieve common objectives. Early explorations, such as HuggingGPT, utilize a LLM as a controller to coordinate multiple models from the Hugging Face Hub to collaboratively complete multimodal tasks \cite{shen2023hugginggptsolvingaitasks}. This can be considered a nascent form of collaborative planning. Multi-agent collaborative planning demonstrates significant potential across various domains. For instance, CodeEdu \cite{zhao2025codeedumultiagentcollaborativeplatform} and AI-Powered Math Tutoring \cite{chudziak2025aipoweredmathtutoringplatform} have established multi-agent collaboration platforms for personalized programming and mathematics education, respectively. In complex technical scenarios like 6G network optimization, the RIDAS framework introduces a multi-agent system comprising Representation-Driven Agents (RDAs) and Intention-Driven Agents (IDAs) to bridge the gap between high-level user intentions and low-level network configurations \cite{ding2025ridasmultiagentframeworkairan}. The Aime framework addresses issues such as rigid plan execution and inefficient communication in multi-agent systems, achieving dynamic and reactive group planning \cite{shi2025aimefullyautonomousmultiagentframework}. The focus of planning shifts from "What should I do?" to "How should we divide tasks and collaborate?" This requires agents not only to plan their actions but also to predict and understand the intentions and behaviors of others, enabling strategic interactions in complex social contexts.

At the L5 level, planning envisions autonomous goal setting and value alignment. In this scenario, unlike in Levels 1–4, agents are not merely executors of plans but also proposers of goals and shapers of the environment. Generative Agents enable agents to plan daily behaviors based on their memory streams, successfully simulating credible human social interactions and demonstrating the potential of planning in social simulation \cite{park2023generativeagentsinteractivesimulacra}. Research on LLMs for Agent-Based Modeling systematically explores the application of LLMs throughout the agent-based modeling (ABM) cycle, from problem formulation to result dissemination, providing insights for simulating complex socio-economic systems \cite{vanhée2025largelanguagemodelsagentbased}. The conceptual work Towards AI Urban Planner views urban planning as a generative AI task, where agents generate land-use plans under various constraints, representing a grand vision of agents participating in transforming the physical world \cite{fu2025urbanplaningaiagent}. The SE-VLN framework introduces self-evolution capabilities, allowing agents to continuously learn and evolve during testing, a key feature toward L5 adaptive systems \cite{dong2025sevlnselfevolvingvisionlanguagenavigation}. Discussions on L5 planning touch upon the ultimate question in artificial intelligence: Can machines have their own vision? This necessitates a deep integration of planning capabilities with value systems, where agents autonomously generate creative goals aligned with long-term interests within a dynamically evolving value framework.

Reliability and Uncertainty Management: The real world is dynamic and unpredictable. LLM-DP  is designed for dynamic interactive environments \cite{dagan2023dynamicplanningllm}. It formalizes feedback into PDDL  and utilizes BFS solvers to adapt to changes.The AgentOps framework, introduced in Taming Uncertainty via Automation, aims to automate the management of intelligent agent systems through observation, analysis, and optimization, enhancing their stability in uncertain environments \cite{moshkovich2025taminguncertaintyautomationobserving}.

Security and Trustworthiness: As agent capabilities increase, new security risks emerge. Logic-layer Prompt Control Injection (LPCI) reveals a novel attack method where attackers embed malicious payloads within memory or tool outputs, enabling delayed or condition-triggered attacks \cite{atta2025logiclayerpromptcontrol}.

Deployment and Implementation: Efficiently deploying complex planning frameworks in real-world environments remains a significant challenge. The Amico framework focuses on building modular, event-driven autonomous agents for embedded systems \cite{yang2025amicoeventdrivenmodularframework}. General Modular Harness for LLM Agents designs universal modular components for gaming environments. AirLLM explores techniques for remotely fine-tuning LLMs in wireless communication scenarios \cite{yang2025airllmdiffusionpolicybasedadaptive}.

Fundamental Theoretical Understanding: The underlying algorithmic mechanisms by which LLMs perform planning are not well understood. The AlgEval framework, proposed in "Position: We Need An Algorithmic Understanding of Generative AI," aims to systematically study the algorithmic primitives learned by LLMs and their combinations, deepening our fundamental understanding of generative AI \cite{eberle2025positionneedalgorithmicunderstanding}.

Deep Integration of Knowledge and Tools: Future planning requires more effective utilization of structured knowledge and external tools. "Advancing Retrieval-Augmented Generation" explores advanced RAG frameworks for enterprise structured data \cite{lewis2021retrievalaugmentedgenerationknowledgeintensivenlp}. KG2data combines knowledge graphs with ReAct agents to support smarter data queries \cite{yao2023reactsynergizingreasoningacting}. Introspection of Thought (INoT) enables LLMs to read and execute code-like dialogue flows, achieving deeper programmatic reasoning \cite{sun2025introspectionthoughthelpsai}.

In summary, the evolution of planning abilities is a critical pathway toward the maturity of industry agents. Future research needs to advance in multiple dimensions, including enhancing planning complexity, ensuring reliability and safety, and deepening fundamental theoretical understanding.

\begin{figure}[!t]
  \centering
  \includegraphics[width=.9\linewidth]{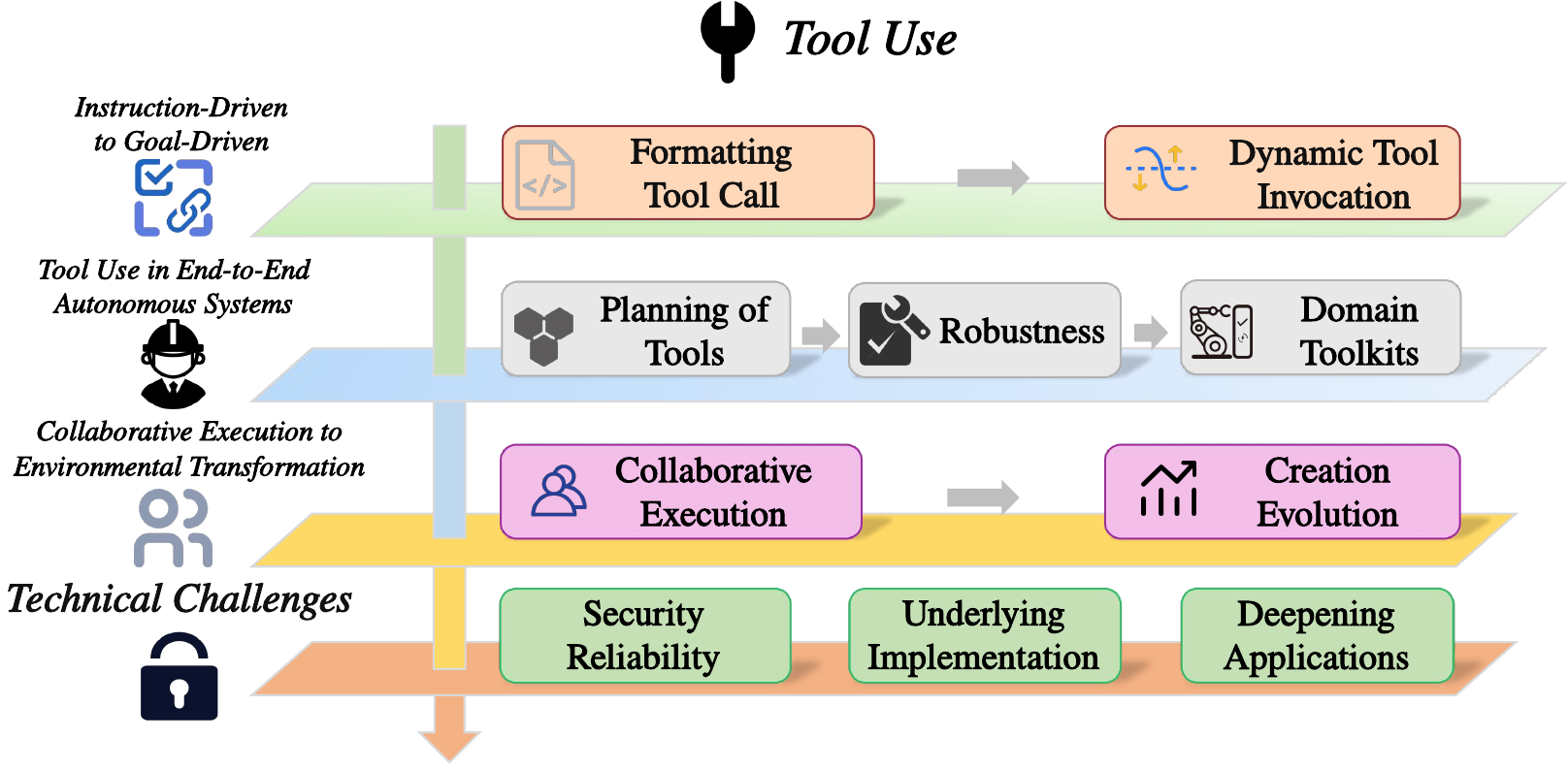}
  \caption{The evolution of tool use in industry agents.}
  \label{fig: tooluse}
\end{figure}

\subsection{Tool Use}
Tool use represents the third core technology distinguishing agents from traditional models. It enables agents to transcend their inherent knowledge and capabilities, facilitating interactions with the expansive digital and physical worlds. For industry agents, tools are pivotal for delving into specific domains, executing specialized tasks, and ensuring the timeliness and accuracy of information. Without tools, agents remain closed, static digital entities; with tools, they evolve into domain experts capable of invoking calculators, accessing databases, browsing the web, controlling software, and even operating physical devices. This section examines how tool use technology has evolved from simple API calls to complex tool creation and analyzes how this progression supports the transformation of industry agents from basic Q\&A systems to complex systems capable of autonomously modifying their environments.

\subsubsection{From Instruction-Driven to Goal-Driven}

At the L1-L2 stages, tool use addresses two fundamental limitations of agents: the latency of factual knowledge and the lack of precise response capabilities. The core transition is from no tools to having a toolbox, marking the shift of agents from mere thinkers to preliminary doers.

L1 tool usage is solidified and implicitly instruction-driven. At this stage, tools function more as inherent capabilities of the model rather than selectable external modules. Their invocation is fixed and non-selective. The success of CoT essentially treats reasoning as an implicit tool \cite{wei2023chainofthoughtpromptingelicitsreasoning}. Program-Aided Language Models (PAL) further advance this by using code interpreters as fixed external tools, generating code to solve mathematical and logical problems, significantly enhancing result determinism \cite{gao2023palprogramaidedlanguagemodels}. These early explorations, along with powerful foundational models like GPT-4 and Claude, lay the groundwork for more complex tool use \cite{openai2024gpt4technicalreport}. In this phase, tool use is passive and predefined; the agent is unaware of its tool usage, following a specific output format, limiting flexibility and generalization.

Level 2 tool usage evolves into goal-driven selective invocation. The agent begins to act as a dispatcher, capable of selecting and invoking appropriate tools from a predefined set based on task requirements. Toolformer serves as a pioneering work in this stage, enabling LLMs to autonomously decide when, where, and how to call APIs through self-supervised learning \cite{schick2023toolformerlanguagemodelsteach}. The ReAct framework provides a core action paradigm by interleaving thought and action, allowing the agent to dynamically interact with tools \cite{yao2023reactsynergizingreasoningacting}. Building on this, the variety and scale of tools expand rapidly: WebGPT \cite{nakano2022webgptbrowserassistedquestionansweringhuman} and WebShop \cite{yao2023webshopscalablerealworldweb} explore using the browser as a tool for question answering and interacting within a simulated shopping website, respectively; TALM \cite{parisi2022talmtoolaugmentedlanguage} fine-tunes models to integrate tool outputs into text generation; while ToolLLM \cite{qin2023toolllmfacilitatinglargelanguage}, Gorilla \cite{patil2023gorillalargelanguagemodel}, and TaskMatrix.AI  \cite{liang2023taskmatrixaicompletingtasksconnecting} aim to enable models to master thousands to millions of real-world APIs. To facilitate the development of such applications, open-source frameworks like LangChain and BMTools have emerged, significantly lowering the barrier to entry. Agents gain the freedom to choose tools, greatly expanding their capabilities. However, they remain “tool users,” with their upper capability bound limited by the predefined tool library. They excel at using existing tools but cannot handle novel problems beyond the available tool set.

\subsubsection{Tool Use in End-to-End Autonomous Systems}

As agents progress to L3, becoming end-to-end systems, they encounter complex tasks that require the collaboration of multiple tools and involve uncertain execution processes. At this stage, tool use capabilities manifest as tool composition planning, failure correction, and even preliminary creativity. The agent's role evolves from being a tool user to a tool orchestrator.

The first is the combination and planning of tools. For complex tasks, a single tool often cannot provide a solution. L3 agents must be capable of combining multiple simple tools into a complex toolchain to accomplish tasks. Frameworks like ART \cite{paranjape2023artautomaticmultistepreasoning} and Chameleon \cite{chameleonteam2025chameleonmixedmodalearlyfusionfoundation} enable agents to perform multi-step reasoning, autonomously decompose tasks, and plan a sequence of tool invocations. ToolChain \cite{zhuang2023toolchainefficientactionspace} and ToolNet \cite{liu2024toolnetconnectinglargelanguage} introduce methods such as A* search and graph structures to assist agents in navigating and planning within vast tool spaces more efficiently. MetaMCP dynamically aggregates multiple MCP services into a unified MCP instance, supporting middleware processing, and functions as a standard MCP server, allowing seamless integration with any MCP client. Tool composition capability is crucial for agents to solve complex problems \cite{hou2025modelcontextprotocolmcp}. It represents a higher level of planning ability, not only planning "what to do" but also "what tools to use," enabling agents to handle systemic tasks beyond the capability of a single tool.

Next is robustness and self-correction during interaction. Tool invocations in the real world often encounter failures, such as unavailable APIs, incorrect parameters, or abnormal returns. L3 agents must possess the ability to handle these anomalies. The CRITIC framework empowers agents to self-validate and correct through interactions with external tools \cite{gou2024criticlargelanguagemodels}. The study "Butterfly Effects in Toolchains" delves into various causes of parameter filling failures in tool invocations, providing insights to enhance interaction reliability \cite{xiong2025butterflyeffectstoolchainscomprehensive}. In the specialized field of software engineering, tools like LibLMFuzz \cite{hardgrove2025liblmfuzzllmaugmentedfuzztarget} and BugScope \cite{guo2025bugscopelearnbugslike} demonstrate how agents utilize toolchains to autonomously analyze binary files, discover, and fix software errors. Self-correction capability renders tool use resilient. Agents transform from fragile executors to engineers capable of troubleshooting and problem-solving, which is vital for deployment in unreliable real-world environments.

Finally, domain-specific specialized toolkits. At this stage, agents begin to be deeply applied in specific industries, and their toolboxes become increasingly specialized. In the field of scientific discovery, tools like LeanTree \cite{kripner2025leantreeacceleratingwhiteboxproof} and ProofCompass \cite{wischermann2025proofcompassenhancingspecializedprovers}, combined with LLMs, accelerate formal theorem proving in environments like Lean 4. In healthcare, the OrthoInsight \cite{wu2025orthoinsightribfracturediagnosis} framework integrates the YOLOv9 model and medical knowledge graphs as tools to assist doctors in diagnosing rib fractures. In code development, ToolCoder \cite{zhang2023toolcoderteachcodegeneration} trains models to use API search engines to discover and utilize unfamiliar APIs, enhancing code generation capabilities. This signifies the shift of tool use from general-purpose to specialized, forming the foundation for industry agents to create core value.

\subsubsection{From Collaborative Execution to Environmental Transformation}

At L4 and above, the focus of tool use shifts from individual capabilities to collective collaboration, ultimately aiming toward the agent's active transformation of its environment.

L4 tool usage is collaborative execution. At this level, multiple agents form a team to collaboratively operate a shared set of tools to achieve grand objectives. HuggingGPT serves as an early example of this concept. It utilizes ChatGPT as a decision-maker to orchestrate various models from the Hugging Face community to address multimodal tasks \cite{shen2023hugginggptsolvingaitasks}. In more specific industry applications, systems like CodeEdu \cite{zhao2025codeedumultiagentcollaborativeplatform} have constructed multi-agent platforms that combine tools to provide personalized programming education to students; GasAgent \cite{zheng2025gasagentmultiagentframeworkautomated} employs multi-agent systems to automatically optimize gas usage in smart contracts; IM-Chat \cite{lee2025imchatmultiagentllmbasedframework} facilitates knowledge transfer in the injection molding industry through a multi-agent framework. To better design and manage such complex workflows, FlowForge offers an interactive visualization tool as a foundational environment for building multi-agent workflows \cite{hao2025flowforgeguidingcreationmultiagent}. In these scenarios, the unit of tool use transitions from individuals to organizations, which not only enhances the scale and complexity of tasks but also introduces new challenges such as resource allocation, task scheduling, and collaborative operations, making tool management itself a complex planning problem.

L5 tool usage involves creation and evolution. This represents the highest form of tool use, where agents are no longer merely users of tools but become creators of tools. Early explorations of autonomous agents, such as Auto-GPT \cite{yang2023autogptonlinedecisionmaking} and BabyAGI \cite{talebirad2023multiagentcollaborationharnessingpower}, autonomously link existing tools to accomplish open-ended goals, demonstrating a nascent form of this autonomy. The CREATOR framework stands as a landmark in this direction \cite{qian2024creatortoolcreationdisentangling}. It allows LLMs to identify capability gaps during problem-solving and autonomously create new tools. This tool creation capability enables agents to transform from mere adaptors of their environment to active modifiers of it. They can dynamically expand their capabilities based on needs, rather than passively waiting for humans to provide new tools or retrieve existing ones. This meta-capability is a crucial step toward true autonomy and general intelligence, though it remains an area requiring further exploration.

\subsubsection{Challenges of Tool Use in Real-World}

The rapid development in the field of tool usage is accompanied by a series of critical challenges spanning evaluation, security, implementation mechanisms, and other aspects, giving rise to extensive frontier research:

Security and Reliability: As agents increasingly connect with real-world APIs, security issues become more prominent. ToolSword systematically exposes security vulnerabilities across the three stages of tool learning: selection, execution, and integration \cite{ye2024toolswordunveilingsafetyissues}. InjecAgent specifically evaluates "indirect prompt injection" attacks against tool-integrated agents \cite{zhan2024injecagentbenchmarkingindirectprompt}.

Underlying Implementation and Optimization: Researchers are also exploring more fundamental implementation mechanisms. ToolkenGPT \cite{hao2024toolkengptaugmentingfrozenlanguage} proposes representing tools as special Toolken embeddings integrated into the model's vocabulary, enabling even non-retrained models to use tools. Probing Information Distribution in Transformer Architectures \cite{buonanno2025probinginformationdistributiontransformer} uses entropy analysis to explore how information flows within the model. Teach Old SAEs New Domain Tricks \cite{koriagin2025teacholdsaesnew} investigates how to adapt models to new domains and tools without full retraining.

Deepening Industry Applications: Deploying tool-enabled agents in real-world industries requires overcoming the challenge of insufficient realism in simulated industry environments. The AgentFly framework aims to enhance the capabilities of language model agents through reinforcement learning \cite{wang2025agentflyextensiblescalablereinforcement}. WebShaper utilizes tools for information retrieval to construct high-quality datasets in a more automated manner \cite{tao2025webshaperagenticallydatasynthesizing}. However, a significant gap remains between existing tool-calling environments and real-world scenarios.

\section{Application Practice of Industry Agents}

After systematically analyzing the three foundational technologies supporting agent capabilities—memory, planning, and tool usage, this chapter shifts focus to the application practices of industry agents. Using the L1 to L5 capability hierarchy framework, we comprehensively review and present the concrete implementations of industry agents in the real world. The evolution of these three technologies is not an isolated theoretical exploration; rather, it is deeply intertwined with the roles agents play across various industries, the complexity of problems they address, and the depth of value they create.

At the L1 level, agents function as Process Execution Systems, accurately translating human instructions. At the L2 level, they evolve into Interactive Problem-Solving Systems, becoming effective assistants to humans. At the L3 level, they operate as End-to-End Autonomous Systems, independently completing complex tasks within a domain. At the L4 level, agents transform into Collaborative Intelligent Systems, shifting focus from individuals to organizations, executing complex business processes, or conducting system simulations through group collaboration. Ultimately, at the L5 level, agents reach the pinnacle as Adaptive Social Systems, not only adapting to the environment but also becoming creators capable of autonomously generating goals, evolving values, and co-evolving with the environment. This chapter analyzes representative research and application cases at each level, depicting a panoramic view of the development of industry agents from theory to practice.

\begin{figure}[!t]
  \centering
  \includegraphics[width=.95\linewidth]{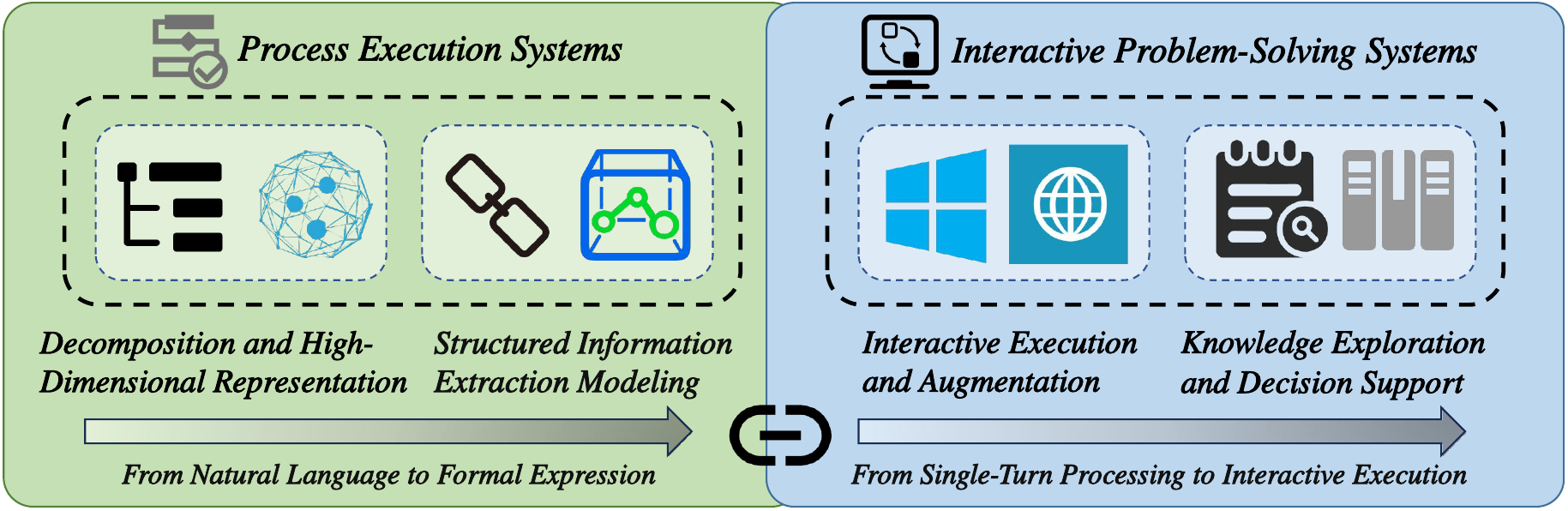}
  \caption{The process execution system and interactive problem-solving system.}
  \label{fig: process}
\end{figure}

\subsection{Process Execution System}

At the L1 level, agents serve as process execution systems. Their core value lies in being reliable extensions of human instructions, primarily manifested in two aspects: accurately translating unstructured human language into machine-executable formal languages, and automating the extraction of structured data from vast amounts of information to execute fixed business rules. At this stage, agents act as fundamental translators and executors in the digital world.

\subsubsection{Translation from Natural Language to Formal Language}

Seamlessly converting natural language into formal language is a key capability of L1 agents, significantly lowering the usage threshold of professional software and systems.

In the field of database interaction, Text-to-SQL technology is a typical representative. HydraNet~\cite{lyu2020hybridrankingnetworktexttosql} innovatively formulates the Text-to-SQL task as a column-wise ranking problem, effectively leveraging the native capabilities of pre-trained LLMs, achieving leading performance on benchmarks like WikiSQL~\cite{zhong2017seq2sqlgeneratingstructuredqueries}. To further enhance the accuracy of complex queries, the SQL-PaLM framework combines few-shot prompting, instruction fine-tuning, and execution feedback mechanisms, significantly improving model performance~\cite{sun2024sqlpalmimprovedlargelanguage}. DIN-SQL~\cite{pourreza2023dinsqldecomposedincontextlearning} adopts a strategy of decomposing complex problems into sub-problems and solving them step by step, achieving new state-of-the-art levels on more challenging benchmarks like Spider~\cite{lei2025spider20evaluatinglanguage} and BIRD \cite{li2023llmservedatabaseinterface}. Addressing specific industry needs, FinStat2SQL designs a lightweight and efficient multi-agent framework tailored to Vietnamese accounting standards, demonstrating the application potential of this technology in specialized fields~\cite{nguyen2025finstat2sqltext2sqlpipelinefinancial}.

In the field of industrial design, Text-to-CAD is another important application direction, aiming to directly convert product descriptions into three-dimensional models. Various technical paths have been proposed: CADFusion~\cite{wang2025texttocadgenerationinfusingvisual} and Text2CAD~\cite{yavartanoo2024text2cadtext3dcad} ensure the geometric accuracy and logical coherence of models through visual feedback and intermediate view generation. Other methods, such as CadQuery~\cite{xie2025texttocadquerynewparadigmcad} and CAD-Coder~\cite{guan2025cadcodertexttocadgenerationchainofthought}, directly generate executable CAD modeling scripts, utilizing the determinism of code to ensure generation quality. To address the scarcity of training data, works like CADmium~\cite{govindarajan2025cadmiumfinetuningcodelanguage} and CAD-Llama~\cite{li2025cadllamaleveraginglargelanguage} significantly enhance model generation capabilities through large-scale dataset generation and adaptive pre-training.

Additionally, FlexCAD~\cite{zhang2025flexcadunifiedversatilecontrollable} achieves controllable generation across construction hierarchies, while CAD-MLLM~\cite{xu2025cadmllmunifyingmultimodalityconditionedcad} constructs the first multimodal CAD framework capable of processing inputs from text, images, or point clouds, demonstrating stronger versatility.

\subsubsection{Structured Information Extraction and Processing}

Beyond language translation, L1 agents are widely applied in extracting key information from unstructured documents and data streams. The LayoutLM series models, including LayoutLM~\cite{xu2020layoutlmpretrainingtextlayout} and LayoutXLM~\cite{xu2021layoutxlmmultimodalpretrainingmultilingual}, represent pioneering work in this field. By jointly modeling text, layout, and visual information, they significantly enhance information extraction accuracy in rich-text documents such as forms and receipts. 

This technology has also given rise to new application paradigms. For instance, an end-to-end framework based on LLMs has been implemented to automate telephone surveys and result analysis, greatly improving data collection efficiency in fields like healthcare~\cite{kaiyrbekov2025automatedsurveycollectionllmbased}. In real-time data processing, LLMs are also utilized to augment traditional machine learning models. For example, in spam detection tasks, ensemble methods significantly enhance the system's robustness and adaptability~\cite{lee2025enhancingphishingemailidentification}.

\subsection{Interactive Problem-Solving System}

When agents evolve to L2, they transition from simple command executors to interactive problem-solving systems, serving as a copilot or assistant to humans in the digital world. Their capabilities are reflected in two core scenarios: first, as efficient tools that enhance human execution through interaction with software and web environments; second, as knowledgeable advisors that improve human decision-making through knowledge exploration and integration.

\subsubsection{Interactive Execution and Augmentation}

One of the core tasks of L2 agents is to understand user intent and translate it into a series of actions on graphical user interfaces (GUIs) or web pages, thereby automating tasks.
In desktop and web application automation, a series of innovative frameworks have emerged. UFO~\cite{zhang2024ufouifocusedagentwindows} and LLMPA~\cite{guan2023intelligentvirtualassistantsllmbased} leverage large visual or language models to enable natural language control of Windows and mobile applications. CogAgent~\cite{hong2024cogagentvisuallanguagemodel} and SeeClick~\cite{cheng2024seeclickharnessingguigrounding} enhance GUI element recognition accuracy through optimized visual encoding and localization pre-training. WebVoyager \cite{he2024webvoyagerbuildingendtoendweb} and WebAgent~\cite{gur2024realworldwebagentplanninglong} focus on web environments, completing complex open-ended tasks on real websites by integrating multimodal information or employing modular program generation. To improve the robustness of these interactions, LASER~\cite{ma2024laserllmagentstatespace} and Rollback~\cite{zhang2025enhancingwebagentsexplicit} mechanisms introduce backtracking capabilities, allowing agents to recover from errors. Algorithms like Language Agent Tree Search and Best-first tree search enhance the success rate of complex tasks through more systematic exploration and planning. WebArena~\cite{zhou2024webarenarealisticwebenvironment} and Mind2Web~\cite{deng2023mind2webgeneralistagentweb} provide evaluation environments that include real websites and diverse tasks. OpenAgents~\cite{xie2023openagentsopenplatformlanguage} and OpenWebAgent~\cite{zhang2025litewebagentopensourcesuitevlmbased} open-source platforms lower development barriers, promoting the application of these technologies in real-world scenarios.

Additionally, extensive research focuses on optimizing data, model architectures, and learning paradigms. ScribeAgent~\cite{shen2024scribeagentspecializedwebagents} and WEPO~\cite{liu2024wepowebelementpreference} enhance model performance by utilizing production-level workflow data and unsupervised preference learning, respectively. Agent-E~\cite{abuelsaad2024agenteautonomouswebnavigation} and R2D2~\cite{huang2025r2d2rememberingreflectingdynamic} achieve more efficient environmental perception and memory utilization through architectural optimization. SkillWeaver~\cite{zheng2025skillweaverwebagentsselfimprove} and ASI~\cite{wang2025inducingprogrammaticskillsagentic} explore methods for agents to autonomously learn and utilize reusable skills, improving their self-improvement capabilities. On mobile devices, Mobile-Agent-v2~\cite{wang2024mobileagentv2mobiledeviceoperation} addresses navigation challenges in mobile operations through a three-agent architecture involving planning, decision-making, and reflection. VisionTasker~\cite{song2024visiontaskermobiletaskautomation} and DroidBot-GPT~\cite{wen2024droidbotgptgptpowereduiautomation} achieve precise mobile task automation by utilizing visual UI understanding and natural language conversion of GUI states.
These efforts collectively form the technological foundation for L2 agents as digital labor forces, liberating humans from tedious repetitive tasks.

\subsubsection{Knowledge Exploration and Decision Support}

As advisors, L2 agents leverage their strong language understanding and tool utilization capabilities to provide in-depth decision support to humans in specialized fields. This capability is grounded in frameworks like ReAct, Toolformer, and PAL, which enable LLMs to call external tools, execute code, and interact with knowledge bases.

In the financial sector, agents are employed for market analysis and strategy simulation. The LLM-Trader framework analyzes market dynamics by simulating interactions of trading agents, while ElliottAgents~\cite{chudziak2025elliottagentsnaturallanguagedrivenmultiagent} constructs multi-agent systems for collaborative technical analysis. Proprietary models like Agentar-Fin-R1~\cite{zheng2025agentarfinr1enhancingfinancialintelligence} are developed to enhance financial intelligence, and the InvestAlign~\cite{wang2025investalignovercomingdatascarcity} framework improves model interpretability by aligning with human decision-making processes.

In the fields of science and medicine, agents are becoming valuable assistants to researchers and doctors. ChemCrow~\cite{bran2023chemcrowaugmentinglargelanguagemodels} and ChemAgent~\cite{tang2025chemagentselfupdatinglibrarylarge} equip LLMs with specialized chemical toolsets, enabling them to autonomously plan and execute tasks like chemical synthesis. In medicine, Discuss-RAG~\cite{dong2025talkretrieveagentleddiscussions} enhances the accuracy of medical question-answering through multi-agent debates~\cite{dong2025talkretrieveagentleddiscussions}, MEDDxAgent~\cite{rose2025meddxagentunifiedmodularagent} improves differential diagnosis through iterative learning, and frameworks like MedRAX~\cite{fallahpour2025medraxmedicalreasoningagent} and MedAide \cite{yang2025medaideinformationfusionanatomy} integrate multi-source medical data to provide reliable support for complex queries.
 
The academic research process itself has also become an object of automation. Benchmarks like ResearchArena~\cite{kang2025researcharenabenchmarkinglargelanguage} and LitSearch~\cite{ajith2024litsearchretrievalbenchmarkscientific} are used to evaluate LLM performance in literature retrieval and review tasks. Systems like CiteAgent~\cite{press2024citemelanguagemodelsaccurately} and ResearchAgent~\cite{baek2025researchagentiterativeresearchidea} can autonomously read papers, attribute citations, or propose new research ideas. Frameworks like Agent Laboratory~\cite{schmidgall2025agentlaboratoryusingllm} and AI Scientist~\cite{lu2024ai, yamada2025aiscientistv2workshoplevelautomated} achieve full-process automation from research conception to paper writing, demonstrating the immense potential of AI in accelerating scientific discovery. 

Furthermore, L2 agents' capabilities are widely applied in more specialized fields, showcasing their potential as empowering platforms. In education, agents are becoming personalized tutors. EduAgent~\cite{xu2024eduagentgenerativestudentagents} simulates student behavior using cognitive priors to generate learning data, while IntelliTutor~\cite{chen2024empoweringprivatetutoringchaining} constructs a comprehensive intelligent tutoring system covering course planning, personalized teaching, and assessment.

In the legal field, the AdvEvol~\cite{chen2025agentcourtsimulatingcourtadversarial} framework enhances legal agents' dynamic knowledge learning and reasoning abilities through adversarial evolution in simulated courtrooms. In content creation, proprietary LLMs like Weaver surpass general models in creative and professional writing tasks through targeted fine-tuning. In code generation, MapCoder~\cite{islam2024mapcodermultiagentcodegeneration} simulates the full cycle of human software development through a multi-agent framework, while StarCoder achieves performance breakthroughs on open-source code large models, reaching levels comparable to many closed-source models.

\begin{figure}[!t]
  \centering
  \includegraphics[width=.9\linewidth]{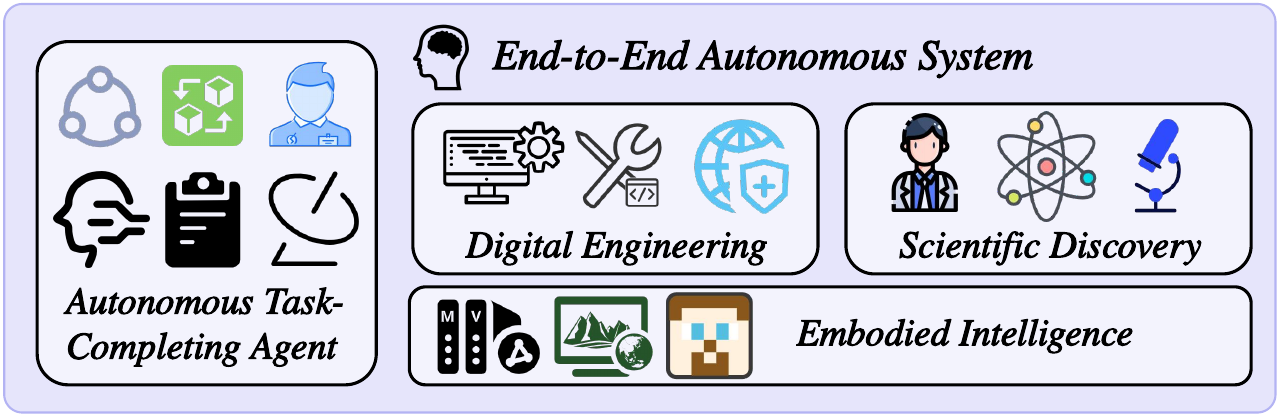}
  \caption{The end-to-end autonomous system.}
  \label{fig: end2end}
\end{figure}

\subsection{End-to-End Autonomous System}
Autonomous systems that reach Level 3 (L3) are end-to-end autonomous agents. These agents are no longer merely assistants to humans; they can autonomously handle the entire process of receiving high-level goals and ultimately completing tasks within a complex domain. Based on their work areas, these systems can be categorized into three major types: the digital world, the physical world, and scientific exploration.

\subsubsection{Autonomous Digital Engineering}
In the digital world, L3 agents are gradually taking on roles such as software engineers, system operation and maintenance experts, and cybersecurity analysts.

In the software engineering field, autonomous agents are capable of automating complex development tasks. Frameworks like AutoDev~\cite{tufano2024autodevautomatedaidrivendevelopment} and SWE-Dev~\cite{du2025swedevevaluatingtrainingautonomous} provide a secure execution environment and high-quality training data for these agents. The Self-Collaboration framework improves the quality of complex code generation by simulating the collaboration patterns of human development teams~\cite{dong2024selfcollaborationcodegenerationchatgpt}. CodePlan~\cite{bairi2023codeplan} and LocAgent~\cite{chen2025locagentgraphguidedllmagents} address challenges in editing and locating repository-level code through incremental dependency analysis and heterogeneous graph representations, respectively. For program repair, RepairAgent~\cite{bouzenia2024repairagentautonomousllmbasedagent} can autonomously fix a large number of errors in the Defects4J dataset, while ChatRepair and ContrastRepair enhance repair efficiency through conversational iterative feedback~\cite{kong2024contrastrepairenhancingconversationbasedautomated}.

In cybersecurity, L3 agents demonstrate the ability to perform automated penetration testing. PentestGPT addresses the context loss problem by using multi-module self-interaction~\cite{deng2024pentestgptllmempoweredautomaticpenetration}. HackSynth~\cite{muzsai2024hacksynthllmagentevaluation} iteratively generates attack instructions through a planner and summarizer, while EnIGMA~\cite{abramovich2025enigmainteractivetoolssubstantially} introduces innovative interactive tools that enable agents to operate complex programs like debuggers, achieving leading performance in CTF benchmark tests.

In system operation and management, L3 agents focus on achieving autonomous diagnosis and recovery of cloud services. Frameworks like RCAgent~\cite{wang2024rcagentcloudrootcause} and TAMO~\cite{wang2025tamofinegrainedrootcauseanalysis} leverage tool-enhanced LLMs to perform root cause analysis for industrial systems or microservices architectures. Systems like AIOpsLab~\cite{shetty2024buildingaiagentsautonomous} and ServiceOdyssey~\cite{yu2025enablingautonomicmicroservicemanagement} enable agents to autonomously manage and repair microservices through simulated fault injection and iterative exploration. Additionally, frameworks like OptiGuide~\cite{li2023largelanguagemodelssupply} and ARS~\cite{li2025arsautomaticroutingsolver} demonstrate the capability of LLM agents to automatically generate efficient heuristic algorithms for complex decision-making problems such as combinatorial optimization. 

\subsubsection{Autonomous Scientific Discovery}

L3 scientific agents go beyond the assisting role of L2, becoming "AI scientists" capable of conducting independent research. These agents can autonomously propose hypotheses, design and execute experiments, analyze data, and ultimately generate new scientific knowledge.

Several general frameworks have explored this grand vision. The Agent Laboratory and AI Scientist frameworks automate the entire process, from research conception to paper publication~\cite{lu2024ai, yamada2025aiscientistv2workshoplevelautomated}. AI Scientist-v2 even generated the first academic paper entirely written by AI and successfully peer-reviewed, marking a milestone achievement~\cite{yamada2025aiscientistv2workshoplevelautomated}. In specific scientific fields, L3 agents have made significant progress. In materials science, frameworks like LLMatDesign~\cite{jia2024llmatdesignautonomousmaterialsdiscovery} and CrystaLLM~\cite{antunes2024crystalstructuregenerationautoregressive} can autonomously design, modify, and evaluate the crystal structures of new materials. In the fields of chemistry and drug discovery, systems such as Coscientist and DrugAssist~\cite{ock2025largelanguagemodelagent} integrate various tools to automate complex experimental workflows or perform end-to-end drug molecule optimization. Frameworks like AgentDrug~\cite{le2025agentdrugutilizinglargelanguage} and LIDDiA~\cite{averly2025liddialanguagebasedintelligentdrug} further enhance molecular optimization accuracy through refinement cycles or intelligent exploration. Robotic systems like ORGANA~\cite{darvish2025organaroboticassistantautomated} and ARChemist~\cite{fakhruldeen2022archemistautonomousroboticchemistry} extend this autonomy into physical laboratories, executing chemists' instructions through controlling robotic arms and experimental equipment, achieving a closed loop of digital intelligence and physical operations.

\subsubsection{Embodied Intelligence}

Embodied intelligence is the third important direction for L3, aiming to give agents the ability to perceive, interact, and learn in the physical world or highly simulated virtual environments. Voyager represents a milestone in this field, achieving lifelong learning in Minecraft without human intervention through automatic curricula, skill libraries, and iterative prompting mechanisms~\cite{wang2023voyageropenendedembodiedagent}. The GITM framework~\cite{zhu2023ghostminecraftgenerallycapable} further improves agents' task completion abilities in virtual worlds by integrating external knowledge. To transfer this capability to real-world robots, researchers focus on aligning language, vision, and actions. PaLM-E is the first work to integrate continuous sensor modalities directly into a language model, enabling end-to-end embodied reasoning~\cite{driess2023palmeembodiedmultimodallanguage}. ECoT and its variants enhance a robot’s generalization ability in complex tasks by introducing multi-step reasoning training~\cite{zawalski2025roboticcontrolembodiedchainofthought}. Frameworks like AdaPlanner~\cite{sun2023adaplanneradaptiveplanningfeedback} and TaPA~\cite{wu2023embodiedtaskplanninglarge} focus on improving agents' planning robustness in dynamic environments, enabling them to adjust plans adaptively based on physical constraints and environmental feedback. These advancements are driving the creation of general-purpose robots capable of autonomously executing tasks in the physical world.

\subsection{Collaborative Intelligent System}

At Level 4, the core capability of intelligent agents evolves from individual autonomy to organizational collaboration. The system consists of multiple specialized agents that communicate and collaborate to achieve large-scale goals that individual agents cannot accomplish. Its value is primarily reflected in two directions: first, as a digital labor force cluster, it directly executes complex end-to-end business processes; second, as a digital laboratory, it simulates the behavior of complex socio-economic systems for reasoning and decision support.

\subsubsection{Collaborative Business Execution}
At the L4 stage, multi-agent systems begin to reshape business processes across various industries.

In the field of intelligent manufacturing, several frameworks have been proposed to achieve flexibility and intelligence in production lines. By simulating structures like leader-follower or hierarchical automation pyramids, multi-agent systems enable dynamic resource scheduling and fault recovery. The MASC framework effectively addresses the dynamic rescheduling challenges in flexible job shop scheduling, while the integration of technologies like digital twins (MADTwin)~\cite{marah2024madtwin} and knowledge graphs allows for more accurate predictive maintenance and production planning. Forward-looking frameworks like DeFACT~\cite{yang2024generative} even explore decentralized autonomous production models based on blockchain.

In the supply chain and logistics domain, multi-agent systems are used to optimize complex coordination problems. For example, intelligent coordination is applied to optimize space allocation and traffic control at roll-on/roll-off terminals, or real-time data processing is used to optimize fleet management at open-pit mines. Frameworks like InvAgent~\cite{quan2025invagentlargelanguagemodel} leverage the zero-shot capability of LLMs to enable adaptive decision-making in inventory management, thus improving the resilience of supply chains.

In the financial services industry, multi-agent collaboration becomes key to improving the complexity and robustness of trading strategies. Frameworks like TradingAgents~\cite{xiao2025tradingagentsmultiagentsllmfinancial} and FinCon~\cite{yu2024fincon} simulate collaboration among different roles within a trading company, such as analysts, strategists, and risk managers, to achieve better trading performance. HedgeAgents~\cite{li2025hedgeagentsbalancedawaremultiagentfinancial} focuses on hedging in volatile markets, while frameworks like MASA~\cite{bougie2025citysimmodelingurbanbehaviors} and MADDQN use multi-agent reinforcement learning to dynamically balance portfolio returns and risks. To systematically evaluate these complex systems, specialized simulation and evaluation platforms like StockSim~\cite{papadakis2025stocksimdualmodeorderlevelsimulator} and FinArena~\cite{xu2025finarenahumanagentcollaborationframework} have emerged, and the TwinMarket~\cite{yang2025twinmarketscalablebehavioralsocial} framework uses LLMs to simulate macroeconomic phenomena.

\subsubsection{Complex System Simulation}
Another key value of L4 agents lies in constructing high-fidelity digital laboratories for simulating and reasoning about the dynamic evolution of complex systems such as human societies, economies, and cities.

In the domain of general simulation frameworks, works like the Simulation Agent Framework~\cite{kleiman2025simulation} and LLM-DT~\cite{xia2024llm} focus on combining the natural language interaction capability of LLMs with the rigor of traditional simulation engines, allowing users to build and validate simulation models in a more intuitive manner. AgentSociety~\cite{piao2025agentsocietylargescalesimulationllmdriven} constructs a large-scale social simulator to study complex social dynamics.

In the fields of transportation and urban development, multi-agent simulations demonstrate significant potential for application. CoMAL~\cite{yao2025comalcollaborativemultiagentlarge} and CoLLMLight~\cite{yuan2025collmlightcooperativelargelanguage} optimize mixed traffic flows or urban traffic signals through agent collaboration. In urban planning, frameworks like CUP simulate interactions and negotiations between roles such as planners and residents to generate and evaluate land use proposals, facilitating more dynamic and human-centered urban development. CitySim performs detailed simulations of individual behaviors, enabling predictions of macro-level urban dynamics~\cite{bougie2025citysim}.

These works provide unprecedented, powerful tools for understanding and managing the increasingly complex systems of modern society.

\begin{figure}[!t]
  \centering
  \includegraphics[width=.9\linewidth]{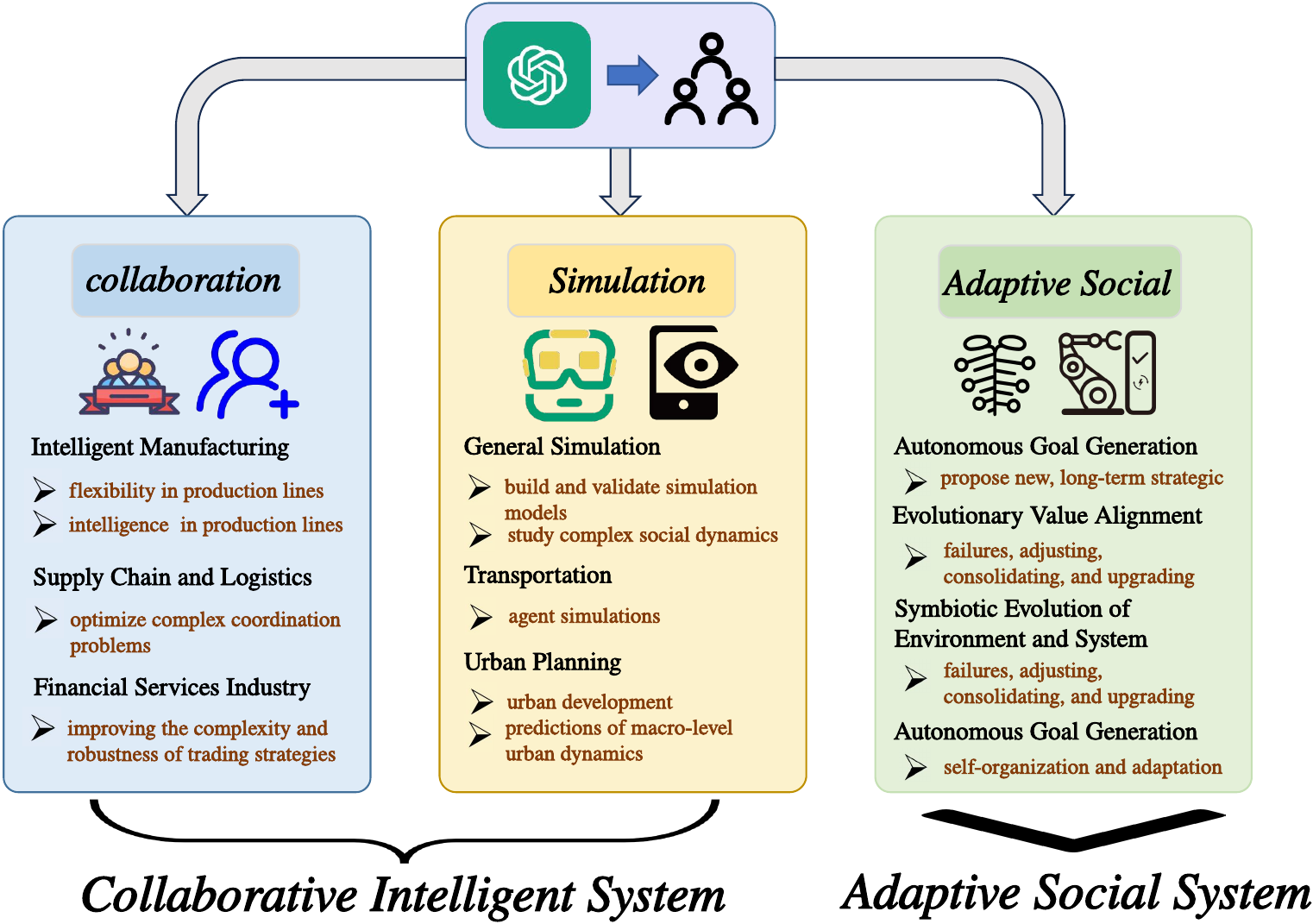}
  \caption{The collaborative intelligent system and adaptive social system.}
  \label{fig: collaborate}
\end{figure}

\subsection{Adaptive Social System}

Level 5 represents the ultimate vision for industry agents—the "adaptive social systems." Unlike L1-L4 systems, which mainly serve as executors of human goals, L5 agents evolve into autonomous entities capable of co-evolving with the environment and human society. These systems do not only adapt to the environment but also actively transform it. They do not simply execute predefined goals but also autonomously generate new objectives and value systems. While no fully realized L5 systems exist yet, their core features are emerging in theoretical discussions and forward-looking research.

The core characteristics of L5 systems can be summarized as follows:

\textbf{Autonomous Goal Generation:} The system no longer passively waits for human input of high-level goals but can autonomously propose new, long-term strategic objectives based on its observations of the environment, internal value systems, and future predictions. Recent explorations in evolutionary agent design, such as EvoAgent~\cite{yuan2025evoagent}, provide early evidence of how agents might autonomously expand their functions and propose novel objectives.

\textbf{Evolutionary Value Alignment:} The values or decision-making criteria of the agent group are not fixed; instead, they evolve through continuous interaction with the environment, learning from collective successes and failures, and adjusting, consolidating, and upgrading over time in a process similar to cultural evolution. This notion resonates with work on evolutionary multi-value alignment in normative multi-agent systems~\cite{riad2023multi}, as well as multi-level frameworks for value alignment in agentic AI systems~\cite{zeng2025multi}.

\textbf{Symbiotic Evolution of Environment and System:} L5 systems actively transform their environments. They change the rules and structures of the physical or digital world through their actions, and these changes, in turn, influence the system’s subsequent development, creating a dynamic, mutually shaping, symbiotic relationship. Adaptive environments such as those modeled in AdaSociety~\cite{huang2024adasociety} highlight the potential for co-evolving system--environment dynamics.

\textbf{Emergent Social Structures:} In the complex interactions among intelligent agents, spontaneous structures, such as organizations, norms, and cultures, emerge that were not explicitly designed, enabling high levels of self-organization and adaptation. Recent research on computational architectures of society and the genesis of social rules~\cite{shan2025computational} illustrates how new forms of norms and institutions can be generated in agent societies.

Although L5 is still at the conceptual stage, its potential applications can be envisioned in several cutting-edge fields. For instance, in the economic domain, an L5 system might manifest as a fully autonomous decentralized organization, where AI agents not only perform business tasks but also formulate company strategies, adjust organizational structures, and even create new business models. In urban governance, an L5 system might go beyond L4’s planning simulations, becoming a city organism that can perceive, decide, and regulate various city resources (e.g., energy, transportation, public services) in real-time, adjusting governance strategies based on the long-term evolution of societal welfare~\cite{piao2025agentsocietylargescalesimulationllmdriven, tang2025gensimgeneralsocialsimulation, gao2025agentscope10developercentricframework, wang2025yulanonesimgenerationsocialsimulator}. In scientific research, an L5 system might form an autonomous scientific community that not only completes specific research (L3/L4) but also proposes new research paradigms, defines important scientific questions, and guides the direction of scientific development.

The challenges of achieving L5 are immense and multidimensional. They involve not only advanced technical problems like complex systems, lifelong learning, and multi-agent game theory but also deep philosophical issues related to control, ethics, and human-machine relationships. However, the exploration of L5 represents our contemplation of the ultimate potential of general artificial intelligence, and its development will profoundly redefine the relationship between humans and intelligence, and technology and society.

\section{Evaluation of Industry Agents}

With the rapid deployment of industry agents in real-world applications, the question of how to evaluate their capabilities in a scientific and comprehensive way has become both crucial and challenging. An effective evaluation system is not only a measure of technological progress but also a foundation for guiding model optimization, ensuring system safety, and building trust across industries.

This section provides a systematic review of evaluation methods and benchmarks for industry agents in practical settings. We begin with three fundamental abilities—memory, planning, and tool usage—and introduce benchmarks designed to assess the general cognitive skills of agents. We then examine specialized evaluation approaches for typical domains such as finance, healthcare, and software engineering, where task requirements are highly specific. Finally, we analyze common and domain-specific challenges in existing evaluation systems, and discuss perspectives for developing the next generation of evaluation frameworks that are more realistic, reliable, and efficient.

\begin{figure}[!t]
  \centering
  \includegraphics[width=.95\linewidth]{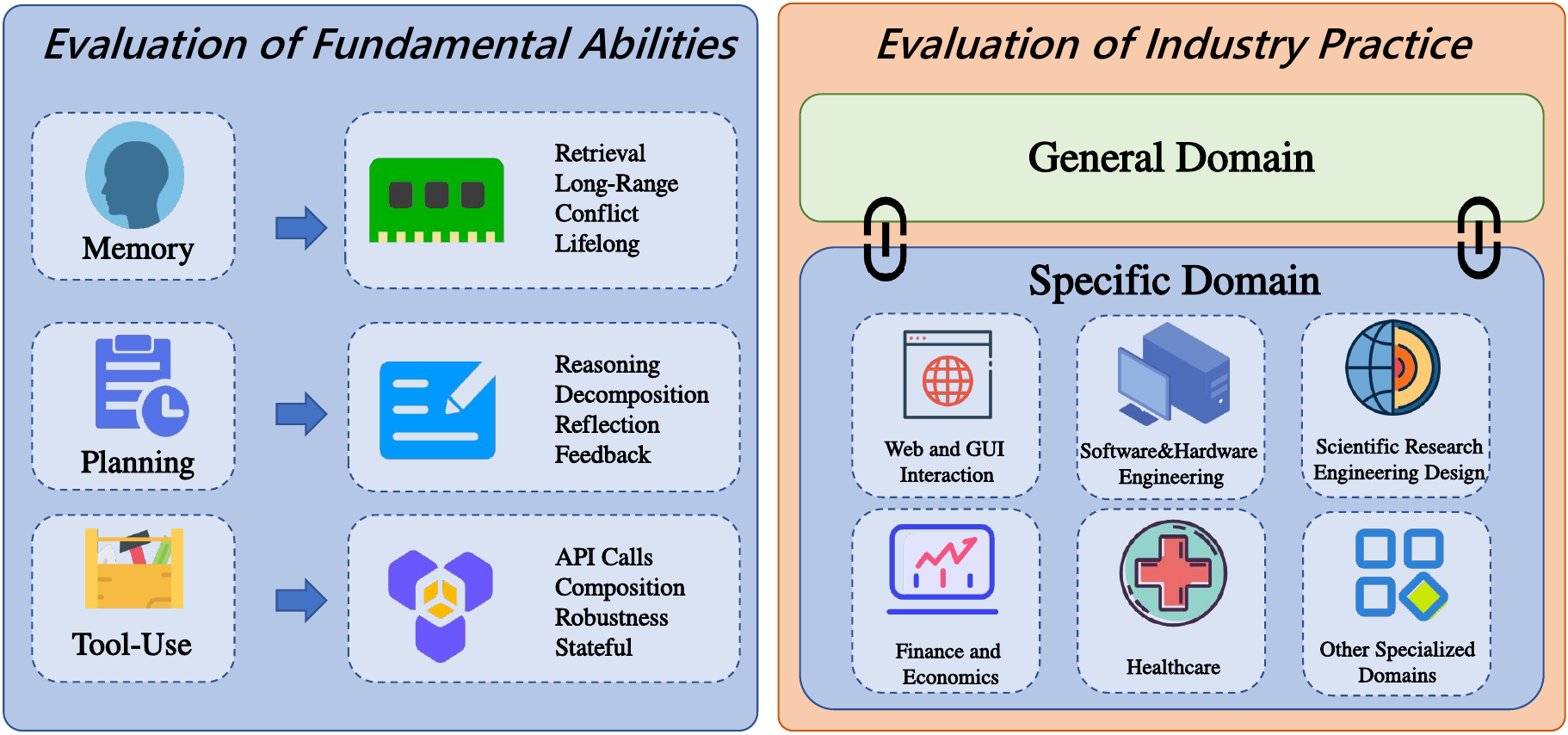}
  \caption{The evaluation of industry agents.}
  \label{fig: evaluation}
\end{figure}

\subsection{Evaluation of Fundamental Abilities}

Before deploying agents in specific industries, it is necessary to conduct a reliable assessment of their underlying cognitive abilities. This section focuses on the evaluation of three fundamental pillars that constitute the core of agents: memory, planning, and tool usage. We discuss how standardized benchmarks and tasks can be used to quantify agent performance in information retention and retrieval, complex task decomposition and execution, and interaction with the external world. The evaluation of these fundamental abilities provides a common language for understanding and comparing different agent architectures, and it serves as a prerequisite for more advanced evaluation in industry-specific applications.

\subsubsection{Evaluation of Memory Abilities}

Memory is the foundation for agents to perform long-term and coherent tasks. Its evaluation focuses on the accuracy of information retrieval, cross-task learning, long-range understanding, and conflict resolution. MemoryAgentBench is a systematic benchmark designed for this purpose\cite{hu2025evaluatingmemoryllmagents}. It specifically evaluates memory agents in four key areas: accurate retrieval, in-test learning, long-range understanding, and conflict resolution. With the expansion of model context windows, the evaluation of long-term memory has become a central topic. The Embodied Long-Context Memory Benchmark introduces 60 embodied tasks in the Habitat simulator that require long-term memory and situational awareness\cite{yadav2025findingdorybenchmarkevaluatememory}. Similarly, 3DMem-Bench provides more than 26,000 trajectories and 2,892 embodied tasks, offering a comprehensive benchmark for assessing long-term memory reasoning in 3D environments\cite{hu20253dllmmemlongtermspatialtemporalmemory}.

In text understanding, the QuALITY \cite{pang2022qualityquestionansweringlong} dataset creates multiple-choice question tasks on long documents with an average length of about 5,000 words, while QMSum \cite{zhong2021qmsumnewbenchmarkquerybased} proposes query-based multi-domain meeting summarization tasks. Together, they challenge models with long-text processing and deep understanding . The LoCoMo \cite{maharana2024evaluatinglongtermconversationalmemory} benchmark further evaluates long-term memory in ultra-long conversations through tasks such as question answering, event summarization, and multimodal dialogue generation. To address these challenges, ReadAgent \cite{lee2024humaninspiredreadingagentgist} introduces an interactive reading mechanism based on memory fragments and key-point memory, achieving significant context expansion in tasks like QuALITY \cite{pang2022qualityquestionansweringlong}. MemGPT employs virtual context management and interruption mechanisms, enabling models to go beyond limited context windows in document analysis and multi-session dialogue. These efforts collectively build evaluation methods from text to embodied tasks, and from single-task scenarios to long-term interactions, thus deepening the understanding of agents' ability to process long-term information.

Beyond long-term memory, lifelong learning and dynamic adaptation are also key evaluation aspects. LifelongAgentBench is the first unified benchmark for systematically evaluating the lifelong learning abilities of LLM-based agents \cite{zheng2025lifelongagentbenchevaluatingllmagents}. It includes skill-based tasks in three interactive environments and provides automatic label verification. StreamBench is an online learning benchmark that focuses on evaluating LLMs in iterative performance improvement under continuous feedback streams\cite{wu2024streambenchbenchmarkingcontinuousimprovement}. A dedicated dynamic dialogue agent evaluation system simulates long-term multi-task interleaved conversations to assess long-term memory, continual learning, and information integration. It reveals new challenges for current large models in natural interactions. These benchmarks extend the scope of evaluation from static knowledge assessment to dynamic learning capabilities.

Furthermore, evaluation frameworks are moving toward more comprehensive and multidimensional systems. MemBench proposes a benchmark that combines fact memory, reflective memory, participatory interactions, and observational scenarios\cite{tan2025membenchcomprehensiveevaluationmemory}. It evaluates the effectiveness, efficiency, and capacity of LLM-based agents’ memory. OST-Bench focuses on incremental observation processing and spatiotemporal reasoning in dynamic exploration tasks. Evaluations also begin to extend into domain-specific applications\cite{lin2025ostbenchevaluatingcapabilitiesmllms}. For instance, the REAL suite is the first benchmark for assessing LLMs in the housing transaction and service domain, covering memory, understanding, and reasoning\cite{zhu2025realbenchmarkingabilitieslarge}. The RAISE framework evaluates agents in real estate sales scenarios with a dual-component memory system and multi-stage evaluation process, showing superior performance over traditional agents in complex multi-turn dialogues\cite{liu2024llmconversationalagentmemory}. Inspired by the Zettelkasten method, some researchers design intelligent memory systems with dynamic indexing and linked knowledge networks, demonstrating their effectiveness across multiple base models. These works significantly enrich memory evaluation methods across different dimensions and application scenarios, making them closer to real-world requirements.

\subsubsection{Evaluation of Planning Abilities}

Planning ability determines the autonomy of agents and sets the upper bound of their problem-solving capacity. Its evaluation covers a wide range of scenarios, from simple reasoning to complex dynamic decision-making. Mathematical and logical reasoning form the foundation of planning, and several classical benchmarks are widely used in this context. GSM8K provides elementary math word problems that require multi-step reasoning\cite{cobbe2021trainingverifierssolvemath}. The Rationale dataset offers algebra problems for evaluating indirect supervision of program learning through natural language reasoning steps\cite{ling2017programinductionrationalegeneration}. HotpotQA evaluates multi-document reasoning through Wikipedia-based question answering\cite{yang2018hotpotqadatasetdiverseexplainable}. The ARC benchmark introduces a scientific corpus and challenging questions to test advanced reasoning and knowledge integration\cite{clark2018thinksolvedquestionanswering}. StrategyQA focuses on problems that require implicit reasoning steps and strategic decomposition, creating a more realistic environment for multi-hop reasoning\cite{geva2021didaristotleuselaptop}. The MATH benchmark provides competition-level math problems, testing the limits of mathematical reasoning in large models\cite{hendrycks2021measuringmathematicalproblemsolving}. Together, these benchmarks form the basis for evaluating logical and mathematical planning abilities.

With the rise of agent-based systems, evaluation has shifted toward more complex interactive and long-horizon decision tasks. The TextAtari benchmark converts Atari game states into textual descriptions, creating nearly 100 tasks to test language agents in decision-making processes lasting up to 100,000 steps\cite{li2025textatari100kframesgame}. Agent-X evaluates visual-centric agents on multi-step deep reasoning tasks in multimodal environments\cite{ashraf2025agentxevaluatingdeepmultimodal}. FlowBench\cite{xiao2024flowbenchrevisitingbenchmarkingworkflowguided}, the first workflow-guided planning benchmark, spans 51 scenarios across six domains, offering multi-format knowledge representation and multi-level evaluation. NATURAL PLAN introduces real-world tasks such as travel planning and meeting scheduling, highlighting the limitations of current models in complex natural language planning\cite{zheng2024naturalplanbenchmarkingllms}. These benchmarks extend planning evaluation from static problem-solving to dynamic and long-term execution.

At the same time, reflection, revision, and feedback learning in planning processes are becoming increasingly important. The Self-Reflection Benchmark demonstrates that iterative reflection mechanisms can significantly improve LLMs in problem-solving tasks\cite{renze2024selfreflectionllmagentseffects}. Reflection-Bench, inspired by cognitive psychology, provides seven tasks to evaluate cognitive abilities in prediction, decision-making, and counterfactual reasoning\cite{li2025reflectionbenchevaluatingepistemicagency}. MINT evaluates continuous performance by simulating multi-round tool use and natural language feedback\cite{wang2024mintevaluatingllmsmultiturn}. AdaPlanner\cite{sun2023adaplanneradaptiveplanningfeedback} applies adaptive planning with environmental feedback loops to test sequential decision-making in environments such as ALFWorld. LLF-Bench offers a diverse platform to evaluate interactive learning from natural language feedback\cite{cheng2023llfbenchbenchmarkinteractivelearning}. The core of these benchmarks lies in assessing how agents learn from failure and adapt during interaction, which is essential for building robust autonomous systems.

Finally, researchers explore more formal and automated approaches to planning evaluation. The PDDL-to-NL framework enables large-scale evaluation of LLMs in PDDL planning tasks\cite{stein2025automatinggenerationpromptsllmbased}, revealing significant performance gaps compared with symbolic planners. ACPBench provides a scalable automated framework with seven reasoning tasks and 13 planning domains described in formal language, supporting systematic assessment of planning and reasoning abilities\cite{kokel2024acpbenchreasoningactionchange}. These efforts contribute to establishing rigorous and quantifiable standards for planning evaluation.

\subsubsection{Evaluation of Tool-Use Abilities}

Tool use is a core extension of agent capabilities. Its evaluation focuses on the accuracy, robustness, and efficiency of selecting, invoking, and composing real-world APIs. The Berkeley Function-Calling Leaderboard (BFCL) introduces the first comprehensive benchmark for assessing function-calling abilities of LLMs. It covers multilingual settings, parallel and multiple calls, and function relevance detection. ToolBench automatically constructs instruction-tuning datasets and, together with the ToolEval evaluator, systematically measures tool-use ability in real API scenarios. Similarly, the ToolAlpaca framework provides thousands of tool-use examples across more than 400 real APIs\cite{tang2023toolalpacageneralizedtoollearning}. API-Bank serves as a pioneering benchmark for tool-augmented LLMs, offering 73 APIs and 314 annotated dialogues to test API planning, retrieval, and execution\cite{li2023apibankcomprehensivebenchmarktoolaugmented}. APIBench provides standardized evaluation for query-based and code-based API recommendation. Collectively, these benchmarks establish the foundation for tool-use evaluation. Recent models such as NexusRaven-V2 even outperform GPT-4 in nested and composite function-calling tasks.

As tool-use scenarios become more complex, benchmarks evolve toward higher precision and greater depth. Seal-Tools introduces strict format constraints and multidimensional metrics for rigorous assessment \cite{wu2024sealtoolsselfinstructtoollearning}. StateEval evaluates sequential API calling through automatically generated test cases \cite{huang2025evaluatingllmssequentialapi}. ComplexFuncBench \cite{zhong2025complexfuncbenchexploringmultistepconstrained} and NESTFUL \cite{basu2025nestfulbenchmarkevaluatingllms} target multi-step, constrained, and nested function-calling scenarios. DICE-BENCH employs a dialogue synthesis framework with tool-dependency graphs and multi-agent roles, generating thousands of high-quality function-calling instances \cite{jang2025dicebenchevaluatingtoolusecapabilities}. The T1 benchmark provides multi-domain, multi-turn dialogue datasets with caching mechanisms, enabling standardized evaluation of dependency handling and dynamic replanning \cite{chakraborty2025t1toolorientedconversationaldataset}. ToolSandbox introduces features such as stateful execution, implicit state dependencies, and a built-in user simulator for dynamic trajectory evaluation \cite{lu2025toolsandboxstatefulconversationalinteractive}. API-BLEND offers a large-scale corpus for training and testing tool-augmented LLMs in realistic API settings \cite{basu2024apiblendcomprehensivecorporatraining}. StableToolBench employs virtual API servers and a stable evaluation system to provide scalable and consistent benchmarks for tool learning \cite{guo2025stabletoolbenchstablelargescalebenchmarking}. Together, these works shift evaluation from isolated API calls to complex, dynamic, and stateful tool-chain execution.

In addition, specialized frameworks and domain-specific benchmarks further enrich tool-use evaluation. ToolEmu simulates tool execution with LLMs, enabling automated safety assessment and risk quantification \cite{ruan2024identifyingriskslmagents}. WebShaper generates high-quality datasets for information search tasks using formal synthesis methods \cite{tao2025webshaperagenticallydatasynthesizing}. The FiReAct pipeline leverages semantic-context retrieval to orchestrate actions across tens of thousands of tools \cite{müller2025semanticcontexttoolorchestration}. PyVision dynamically generates and executes Python-based tools, significantly improving multimodal reasoning in vision benchmarks \cite{zhao2025pyvisionagenticvisiondynamic}. AgentDistill transfers structured task-solving modules distilled from teacher agents, enabling efficient knowledge reuse without retraining \cite{qiu2025agentdistilltrainingfreeagentdistillation}. In specialized domains, the CVDP benchmark \cite{pinckney2025comprehensiveverilogdesignproblems} covers 13 categories in hardware design and verification, while RestBench provides high-quality benchmarks with real-world scenarios and gold-standard solution paths for evaluating agents such as RestGPT \cite{song2023restgptconnectinglargelanguage}. These efforts push tool-use evaluation toward more realistic, complex, safe, and domain-oriented directions.

\subsection{Evaluation of Industry Practice}

When agents are applied to specific industries, basic ability evaluation alone is not sufficient. The success of industry applications depends on whether agents can understand and follow complex business logic, leverage domain-specific knowledge, and handle industry-specific risk scenarios. Building specialized benchmarks for each domain is therefore essential. These benchmarks must not only simulate real business processes and data environments but also account for unique challenges, such as the high risk in finance and strict compliance requirements in healthcare.

This section reviews representative benchmarks and methods that have emerged in key industries, including web interaction, software and hardware engineering, finance, healthcare, and scientific research. The goal is to show how evaluation systems evolve from general-purpose testing to domain-oriented assessment, providing a more realistic measure of the practical value of industry agents.

\subsubsection{General Domain}

Before moving into domain-specific applications, a set of benchmarks has been developed to evaluate agents in cross-industry scenarios and complex challenges. The GAIA benchmark includes 466 real-world problems to test reasoning, multimodal processing, and tool use, providing a reference for comparing human and AI performance \cite{mialon2023gaiabenchmarkgeneralai}. AgentBench offers a multi-dimensional evolving benchmark to assess reasoning and decision-making in multi-turn open-ended environments \cite{liu2023agentbenchevaluatingllmsagents}. OSWorld provides a scalable real-computer environment supporting cross-operating system, multimodal task execution and evaluation. To simulate real business operations, TheAgentCompany builds a scalable framework for assessing AI agents in tasks such as web browsing and coding within a software company setting \cite{xu2025theagentcompanybenchmarkingllmagents}. Similarly, CRMArena introduces nine customer service tasks across three roles to measure performance in real CRM scenarios \cite{huang2025crmarenaunderstandingcapacityllm}. Multi-agent frameworks such as Aime \cite{shi2025aimefullyautonomousmultiagentframework} and AgentOrchestra also demonstrate strong task completion and adaptability on GAIA and related benchmarks \cite{zhang2025agentorchestrahierarchicalmultiagentframework}.

Beyond general capabilities, benchmarks target specific common challenges. In safety, RAS-Eval evaluates LLM agents across 11 CWE security vulnerabilities using 80 test cases and 3,802 attack tasks in simulated and real tool-execution environments \cite{fu2025rasevalcomprehensivebenchmarksecurity}. In process adherence, $\tau$-bench tests agents in dynamic user–agent dialogues to measure rule-following and behavioral consistency \cite{yao2024taubenchbenchmarktoolagentuserinteraction}. Its successor, $\tau^2$-bench, extends evaluation fidelity by adding dual-control telecom settings and composite task generation \cite{barres2025tau2benchevaluatingconversationalagents}. For complex collaboration, CREW-Wildfire generates large-scale wildfire response scenarios with heterogeneous agents, maps, and uncertainty to test coordination, communication, and long-term planning \cite{hyun2025crewwildfirebenchmarkingagenticmultiagent}. SOP-Bench covers 10 industrial domains and over 1,800 tasks, measuring planning, reasoning, and tool use in complex standard operating procedures \cite{nandi2025sopbenchcomplexindustrialsops}.

Additional benchmarks address broader general capabilities. HeuriGym provides an open-source framework for generating heuristics in combinatorial optimization \cite{chen2025heurigymagenticbenchmarkllmcrafted}. AssetOpsBench offers a unified environment for Industry 4.0 development and evaluation \cite{patel2025assetopsbenchbenchmarkingaiagents}. The MAPS suite translates multiple benchmarks into 11 languages, enabling standardized multilingual evaluation of agent performance and safety \cite{hofman2025mapsmultilingualbenchmarkglobal}. EconWebArena assesses autonomous agents in multimodal economic tasks on real web platforms \cite{liu2025econwebarenabenchmarkingautonomousagents}. TextAtari converts Atari game states into text to test long-horizon decision-making \cite{li2025textatari100kframesgame}. AmbiK provides a kitchen environment with ambiguous instructions to compare ambiguity-handling methods \cite{ivanova2025ambikdatasetambiguoustasks}. Agent-X evaluates vision-centric agents on multi-step reasoning in real multimodal environments \cite{ashraf2025agentxevaluatingdeepmultimodal}. Agent-RewardBench tests reward modeling for multimodal LLMs across perception, planning, and safety \cite{men2025agentrewardbenchunifiedbenchmarkreward}. IntellAgent is an open-source multi-agent framework that generates diverse synthetic benchmarks for dialogue AI evaluation \cite{levi2025intellagentmultiagentframeworkevaluating}. The Factorio Learning Environment (FLE) uses a game-based setting to measure long-term planning, program synthesis, and resource optimization. STEPS evaluates sequential reasoning in task execution \cite{wang2023stepsbenchmarkorderreasoning}. The HAL~\cite{hal} framework standardizes evaluation across benchmarks, supporting parallel testing and cost tracking. Spider2-V~\cite{cao2024spider2vfarmultimodalagents} introduces the first multimodal agent benchmark focused on professional data science and engineering workflows, featuring 494 real-world tasks to evaluate an agent's ability to automate data workflows through code generation and GUI operations.

Together, these works establish the foundation for measuring whether agents meet the entry requirements for industry applications, bridging general cognitive abilities with practical business readiness.

\subsubsection{Specific Domain}

\paragraph{Web and GUI Interaction}

Webpages and GUIs are the main entry points for agents to interact with the digital world. Benchmarks in this area aim to evaluate automation abilities in real and dynamic web environments. WebArena \cite{zhou2024webarenarealisticwebenvironment} provides a highly realistic and reproducible environment to assess the functional correctness of language-guided agents on complex, long-horizon tasks. VisualWebArena extends this to vision-based multimodal tasks, focusing on the performance of multimodal web agents \cite{koh2024visualwebarenaevaluatingmultimodalagents}. WebVoyager integrates 15 real-world website tasks and establishes an automated evaluation protocol using GPT-4V, offering a reliable standard for open-web agents~\cite{he2024webvoyagerbuildingendtoendweb}. WEBLINX supports large-scale evaluation for conversational web navigation tasks, with 100K interactions and 2,300 expert demonstrations \cite{lù2024weblinxrealworldwebsitenavigation}.

Some benchmarks focus on enterprise-level systems. WorkArena \cite{drouin2024workarenacapablewebagents} and the BrowserGym environment target LLM-based agents in enterprise software, providing frameworks to evaluate task automation in business applications. World of Bits applies workflow-guided exploration to assess the sample efficiency and performance of deep reinforcement learning agents on web tasks \cite{liu2018reinforcementlearningwebinterfaces}. WebShop creates a simulated e-commerce environment with millions of real products and tens of thousands of instructions, evaluating capabilities in compositional instruction following and query reformulation \cite{yao2023webshopscalablerealworldweb}.

As task complexity increases, new benchmarks introduce richer settings. MMInA evaluates embodied agents in real and dynamic environments through multi-hop and multimodal web tasks \cite{tian2025mminabenchmarkingmultihopmultimodal}. WebCanvas \cite{pan2024webcanvasbenchmarkingwebagents} proposes an online evaluation method for dynamic web interaction, including the Mind2Web-Live dataset and novel evaluation metrics. AssistantBench offers 214 automatically evaluated real-world tasks and highlights current model limitations in open-web navigation \cite{yoran2024assistantbenchwebagentssolve}. Security also becomes a major concern. ST-WebAgentBench includes 222 enterprise tasks, six safety and trustworthiness dimensions, and a specialized framework to expose significant vulnerabilities in existing web agents \cite{levy2025stwebagentbenchbenchmarkevaluatingsafety}.

Together, these benchmarks advance the evaluation of web agents from simple page-level interaction toward comprehensive assessment of complex, dynamic, secure, and multimodal capabilities.

\paragraph{Software and Hardware Engineering}

In software engineering, evaluation focuses on measuring an agent’s ability to understand, generate, modify, and repair complex code. SWE-bench is a milestone benchmark \cite{jimenez2024swebenchlanguagemodelsresolve}. It contains 2,294 real GitHub issues with corresponding pull requests and evaluates LLMs on complex code modification tasks. Several variants are derived from it. To reduce evaluation cost, SWE-bench Lite provides a smaller set with 300 tasks. SWE-bench Multimodal (SWE-bench M) includes 617 multimodal task instances and is designed to assess autonomous systems on visual JavaScript software repair \cite{yang2024swebenchmultimodalaisystems}. SWE-PolyBench \cite{rashid2025swepolybenchmultilanguagebenchmarkrepository} proposes a benchmark of 2,110 multilingual instances, supporting repository-level execution evaluation for Java, JavaScript, TypeScript, and Python. However, some studies point out quality issues in SWE-bench \cite{jimenez2024swebenchlanguagemodelsresolve}, such as solution leakage and insufficient test cases, which may cause significant bias in performance evaluation.

In addition to code repair, other tasks also gain attention. HumanEval evaluates functional correctness of programs synthesized from docstrings \cite{chen2021evaluatinglargelanguagemodels}. TDD-Bench Verified provides a high-quality benchmark with 449 real GitHub issues to evaluate automated test generation in test-driven development \cite{ahmed2024tddbenchverifiedllmsgenerate}. ITBench introduces a systematic framework for evaluating AI agents in IT automation tasks \cite{jha2025itbenchevaluatingaiagents}. SWE-Lancer includes more than 1,400 freelance software engineering tasks, covering both independent engineering and management tasks \cite{miserendino2025swelancerfrontierllmsearn}.

Evaluation frameworks also emerge for specialized scenarios. The LLM-based critics framework uses reference-aware intermediate evaluators to automate assessment of code patch executability and semantics on SWE-bench \cite{jimenez2024swebenchlanguagemodelsresolve}. LLM-BSCVM benchmarks vulnerability detection on curated datasets and achieves more than 91\% accuracy \cite{jin2025llmbscvmllmbasedblockchainsmart}. CSR-Bench proposes a benchmark for computer science research projects, evaluating deployment accuracy, efficiency, and related metrics for automated code deployment \cite{xiao2025csrbenchbenchmarkingllmagents}.
\paragraph{Finance and Economics}

The financial domain is characterized by high risk and strict timeliness, which makes the evaluation of intelligent agents especially rigorous. The FinRL Contests series provides standardized benchmarks that cover diverse financial tasks such as stock trading and order execution \cite{wang2025finrlcontestsbenchmarkingdatadriven}. These benchmarks include real-time, high-quality datasets and realistic market environments. The DeepFund real-time fund benchmark tool connects to live stock market data through a multi-agent architecture \cite{li2025llmsprofessionalfundinvestment}. It evaluates the investment performance of several mainstream LLMs under real market conditions without information leakage. The FinArena framework combines multimodal financial data analysis with user interaction to evaluate agents in stock trend prediction and trading simulation \cite{xu2025finarenahumanagentcollaborationframework}. The Agent Trading Arena simulates a zero-sum virtual economy to assess differences in how LLMs handle text and visual data in numerical reasoning tasks. The FINSABER framework performs long-term cross-market backtesting and reveals significant weaknesses in the generalization and robustness of LLM-based timing strategies \cite{li2025llmbasedfinancialinvestingstrategies}.

Beyond trading performance, evaluation of financial knowledge and comprehensive abilities is also essential. FinEval \cite{guo2024finevalchinesefinancialdomain} is a large benchmark with 8,351 questions across four domains: financial academia, industry, security, and agent capabilities. InvestorBench \cite{li2024investorbenchbenchmarkfinancialdecisionmaking} is the first benchmark designed to assess LLM-based agents across different financial decision-making scenarios. FinResearchBench introduces a logic-tree-based Agent-as-a-Judge framework to automatically evaluate financial research agents on seven key task categories \cite{sun2025finresearchbenchlogictreebased}. The FinLLM leaderboard and the Korean-language Won benchmark provide open leaderboards for evaluating the overall performance of financial LLMs \cite{lin2025openfinllmleaderboardfinancial}.

Risk assessment is a distinctive feature of financial evaluation. The Risk-Engineering Framework proposes a three-level stress testing method for financial LLM agents, emphasizing risk metrics at the model, workflow, and system levels \cite{chen2025standardbenchmarksfail}. Meanwhile, EconWebArena focuses on evaluating agents in performing complex economic tasks within real web environments \cite{liu2025econwebarenabenchmarkingautonomousagents}. The Instruct2DS benchmark supports real-time web data collection in domains such as finance, establishing a new standard for the evaluation of automated data collection systems.

Together, these efforts build a comprehensive evaluation system for financial agents, covering trading strategies, knowledge understanding, and risk control.

\paragraph{Healthcare}

Healthcare is a domain with broad application potential but extremely high risks. Benchmarks in this area are characterized by wide coverage, strong specialization, and a focus on safety. Several comprehensive benchmarks are developed to evaluate agent performance in complex clinical settings. MedAgentBoard provides a systematic benchmark to assess multi-agent collaboration, single LLMs, and traditional approaches \cite{zhu2025medagentboardbenchmarkingmultiagentcollaboration}. ClinicalAgent Bench (CAB) covers five clinical dimensions and 18 tasks for comprehensive evaluation \cite{liao2025reflectoolreflectionawaretoolaugmentedclinical}. MedAgentsBench \cite{tang2025medagentsbenchbenchmarkingthinkingmodels} focuses on challenging problems such as complex medical reasoning, diagnosis, and treatment planning. AgentClinic is a multimodal benchmark that evaluates LLMs in simulated clinical environments, including patient interaction, multimodal data collection, and tool use \cite{schmidgall2025agentclinicmultimodalagentbenchmark}. The MedChain benchmark introduces 12,163 personalized, interactive, and sequential clinical cases, offering a comprehensive testbed for LLMs in realistic decision-making \cite{liu2024medchainbridginggapllm}. The DynamiCare framework establishes the first benchmark for dynamic clinical decision-making, built on the MIMIC-Patient dataset.

For specific medical tasks, ReasonMed is the largest dataset for medical reasoning, providing a new standard for training and evaluating medical QA models \cite{sun2025reasonmed370kmultiagentgenerated}. The MEDDx benchmark offers a complete diagnostic evaluation framework supporting iterative diagnostic strategies. SYNUR and SIMORD are the first open-source datasets for nursing observation extraction and medical instruction extraction, introducing new benchmarks for these tasks \cite{corbeil2025empoweringhealthcarepractitionerslanguage}. The CalcQA benchmark evaluates LLMs in clinical calculation scenarios through 100 case-calculator pairs and 281 medical tools \cite{zhu2025mentibridgingmedicalcalculator}. IPDS provides a large dataset of 51,274 cases to evaluate decision support for inpatient care pathways \cite{chen2025mapevaluationmultiagentenhancement}.

Safety and fairness are core concerns in healthcare evaluation. MedSentry offers a benchmark with 5,000 adversarial medical prompts to test multi-agent topologies for safety and defense mechanisms \cite{chen2025medsentryunderstandingmitigatingsafety}. The AMQA benchmark systematically evaluates population bias in LLMs for medical diagnosis using adversarial QA datasets \cite{xiao2025amqaadversarialdatasetbenchmarking}. The Cancer-Myth benchmark reveals serious weaknesses in state-of-the-art LLMs in identifying and correcting flawed premises in cancer-related questions \cite{zhu2025cancermythevaluatingaichatbot}.

Numerous specialized benchmarks also emerge in specific domains. These include ChestAgentBench \cite{fallahpour2025medraxmedicalreasoningagent} and CheXagent \cite{sharma2024cxragentvisionlanguagemodelschest} for radiology image interpretation , Eyecare-Bench  \cite{li2025eyecaregptboostingcomprehensiveophthalmology}and multilingual CLARA \cite{restrepo2024multiophthalinguamultilingualbenchmarkassessing} for ophthalmology, MedAgentGYM \cite{xu2025medagentgymtrainingllmagents} for biomedical coding reasoning, the Echocardiogram QA Dataset for cardiology, and benchmarks for rare disease gene prioritization based on HPO classification and multi-agent evaluation methods \cite{moukheiber2025echoqalargecollectioninstruction}. WSI-Agents integrates expert agents to achieve superior performance across multimodal whole-slide image (WSI) benchmarks \cite{lyu2025wsiagentscollaborativemultiagentmultimodal}. 3MDBench is an open-source framework for simulating and evaluating telemedicine consultations driven by large vision-language models \cite{sviridov20253mdbenchmedicalmultimodalmultiagent}. M3Bench benchmarks automated medical imaging machine learning \cite{feng2025m3buildermultiagentautomatedmachine}. MedDev-Bench introduces an international dataset to evaluate automated systems for regulatory compliance of medical devices \cite{han2025standardapplicabilityjudgmentcrossjurisdictional}. BeNYfits provides a new benchmark for determining user eligibility across overlapping social benefits \cite{rose2025meddxagentunifiedmodularagent}.

Together, these benchmarks form a foundation for developing trustworthy and reliable medical agents by advancing evaluation from general clinical reasoning to specialized diagnostics, safety, and regulatory compliance.

\paragraph{Scientific Research and Engineering Design}

In scientific discovery and engineering design, evaluation aims to measure the potential of agents as “AI scientists” or “AI engineers.” Several general-purpose scientific benchmarks have been proposed. ScienceQA includes about 21K multimodal multiple-choice questions \cite{lu2022learnexplainmultimodalreasoning}. ScienceWorld simulates an interactive text environment to test scientific reasoning \cite{wang2022scienceworldagentsmarter5th}. DISCOVERYWORLD provides a virtual environment for developing and evaluating agents with end-to-end scientific reasoning abilities \cite{jansen2024discoveryworldvirtualenvironmentdeveloping}. Benchmarks such as AAAR-1.0 \cite{lou2025aaar10assessingaispotential}, ScienceAgentBench \cite{chen2025scienceagentbenchrigorousassessmentlanguage}, and CORE-Bench \cite{siegel2024corebenchfosteringcredibilitypublished} focus on professional research tasks, including experimental design, weakness identification in papers, and computational reproducibility . RExBench evaluates the ability to implement research experiments \cite{edwards2025rexbenchcodingagentsautonomously}. SurveyScope provides a standardized benchmark for automatic scientific literature review across 11 computer science domains \cite{shi2025scisagemultiagentframeworkhighquality}. QASPER offers a benchmark for information-seeking question answering over academic documents \cite{dasigi2021datasetinformationseekingquestionsanswers}. The SUPER benchmark is the first to evaluate the capability of LLMs to configure and execute research repository tasks \cite{bogin2024superevaluatingagentssetting}.

Domain-specific benchmarks are also emerging. DrafterBench evaluates revision of civil engineering drawings \cite{li2025drafterbenchbenchmarkinglargelanguage}. FEABench measures the ability of LLMs and their agents to solve problems in physics, mathematics, and engineering using finite element analysis \cite{mudur2025feabenchevaluatinglanguagemodels}. ChemGraph evaluates LLMs of different scales on 13 tasks in automated computational chemistry workflows \cite{pham2025chemgraphagenticframeworkcomputational}. The TopoMAS framework demonstrates efficiency and accuracy in topological materials discovery through comprehensive benchmarking \cite{zhang2025topomaslargelanguagemodel}. AstroMLab-1 shows that AstroSage-70B outperforms both open-source and closed-source models on astronomy tasks \cite{dehaan2025astromlab4benchmarktoppingperformance}. The DREAMS framework achieves less than 1\% average error in the Sol27LC lattice constant benchmark \cite{wang2025dreamsdensityfunctionaltheory}. The Design Agents framework validates efficiency improvements in automotive design processes using industry-standard benchmarks \cite{elrefaie2025aiagentsengineeringdesign}. ThinkGeo evaluates tool use and multi-step planning of LLMs in remote sensing through structured agent tasks \cite{shabbir2025thinkgeoevaluatingtoolaugmentedagents}. GeoMap-Bench is the first benchmark to assess multimodal LLMs in geological map understanding. PhysGym provides a benchmark suite and simulation platform to evaluate scientific reasoning of LLM-based agents in interactive physics environments \cite{chen2025physgymbenchmarkingllmsinteractive}.

Evaluation of data science and interdisciplinary methodologies also receives increasing attention. DataSciBench is a comprehensive benchmark for LLM abilities in data science \cite{zhang2025datascibenchllmagentbenchmark}. The DSMentor \cite{wang2025dsmentorenhancingdatascience} framework demonstrates performance gains for LLM agents on data science tasks using the DSEval and QRData benchmarks. AutoMind achieves superior results over state-of-the-art baselines on two automated data science benchmarks \cite{ou2025automindadaptiveknowledgeableagent}. BIASINSPECTOR provides a benchmark for systematic evaluation of bias detection in structured data by LLM agents \cite{li2025biasinspectordetectingbiasstructured}.Auto-Bench applies causal discovery principles to evaluate scientific discovery in both natural and social sciences \cite{chen2025autobenchautomatedbenchmarkscientific}. NLP4LP introduces a new benchmark dataset for linear programming and mixed-integer linear programming problems \cite{ahmaditeshnizi2023optimusoptimizationmodelingusing}. LAB-Bench includes over 2,400 multiple-choice questions to assess practical capabilities of AI systems in biological research \cite{laurent2024labbenchmeasuringcapabilitieslanguage}. MLGym-Bench covers 13 diverse AI research tasks to evaluate real-world research skills of LLM agents \cite{nathani2025mlgymnewframeworkbenchmark}. SciCode decomposes 80 scientific problems into 338 sub-tasks to evaluate knowledge recall, reasoning, and synthesis in scientific code generation \cite{tian2024scicoderesearchcodingbenchmark}. Finally, the ToolMaker benchmark contains 15 complex computational tasks across domains, assessing the correctness and robustness of tool generation \cite{wölflein2025llmagentsmakingagent}.

\paragraph{Other Specialized Domains}

Beyond the mainstream areas discussed above, evaluation benchmarks are expanding into many specialized domains. In the legal domain, the eSapiens \cite{shi2025esapiensplatformsecureauditable} platform validates improvements in factual consistency through legal corpus retrieval benchmarks and generation quality tests. In linguistics, LingBench++ introduces a framework that integrates structured reasoning traces, step-by-step evaluation protocols, and metadata across more than 90 languages to assess LLMs on complex linguistic tasks \cite{lian2025lingbenchlinguisticallyinformedbenchmarkreasoning}.

In the education domain, LLM-EduBench provides a systematic framework and datasets for evaluating LLM-based agents in subject teaching and professional development \cite{chu2025llmagentseducationadvances}. PBLBench introduces the first free-output and rigorously human-verified benchmark for project-based learning. It applies structured evaluation standards derived from expert analytic hierarchy processes to test multimodal LLMs in complex reasoning and long-context understanding.

In the humanities and social sciences, HSSBench is a multilingual benchmark designed to evaluate multimodal LLMs on interdisciplinary reasoning and knowledge integration \cite{kang2025hssbenchbenchmarkinghumanitiessocial}. In the gaming and simulation domain, VGC-Bench provides a benchmark platform to evaluate multi-agent strategy generalization in the Pokémon battle environment \cite{angliss2025benchmarkgeneralizingdiverseteam}. The Decrypto benchmark adopts a gamified interaction design to fill the gap in evaluating theory-of-mind reasoning in multi-agent systems.

Together, these benchmarks significantly extend the scope of agent evaluation, laying the groundwork for applications and assessment of intelligent agents across broader areas of human knowledge.

\subsection{Limitations of Existing Evaluation Benchmarks}

Although benchmarks and evaluation methods have made significant progress, current systems still face a series of challenges in advancing intelligent agents toward reliable and general industrial applications.

Trade-off between realism and reproducibility: The closer an environment is to the real world, the more randomness and dynamism it contains, making strict reproducibility extremely difficult. In contrast, highly deterministic simulation environments guarantee reproducibility but often diverge from real scenarios. As a result, agents that perform well in simulations may fail in practice.

Contradiction between evaluation cost and efficiency: For complex tasks, especially those involving open-ended responses and multi-step operations, high-quality human evaluation remains the “gold standard.” However, it is costly and time-consuming. Using more powerful LLMs as evaluators can improve efficiency, but issues such as bias, inconsistency, and hallucination of new knowledge introduce new uncertainties into the reliability of results.

Performance overhead of secure sandboxes: To safely evaluate agents that interact with the external world, they must be placed in isolated sandbox environments. Stronger isolation mechanisms, such as independent virtual machines or containers, often introduce significant performance overhead, which reduces evaluation efficiency. Balancing absolute security with high-performance evaluation remains a critical engineering challenge.

High knowledge barriers and timeliness: Domains such as finance, healthcare, and law require not only highly complex expertise but also constant updates, including new financial instruments, clinical guidelines, and regulations. Current benchmarks cannot maintain knowledge bases that are fully synchronized with the latest industry developments. This leads to an inherent lag in evaluating the timeliness of agent knowledge.

Constraints of data privacy and compliance: Real-world industry data and system interfaces, such as bank transaction records, electronic health records, and legal case files, are protected by strict privacy and data protection regulations \cite{gdpr, hipaa}. This makes it difficult for researchers to access high-quality real data for constructing evaluation environments. Relying on synthetic or heavily anonymized data risks losing subtle but critical details, which reduces evaluation validity.

\section{Discussion}

Although the evolutionary path from L1 to L5 clearly demonstrates the leaps in the capabilities of industry agents, a series of profound challenges emerge in translating these technological blueprints into reliable, general-purpose, and beneficial societal productivity \cite{lim2024largelanguagemodelenabledmultiagent}. These challenges are not only related to technical implementation but also touch upon fundamental issues such as the nature of knowledge, the composition of intelligence, and the evolution of systems. This section distills five core deep problems that represent key bottlenecks for current and future industry agents in moving from being usable to being trustworthy and generalizable.

\subsection{The Gap Between Knowledge and Experience}

A significant phenomenon emerges when examining the application practices from L1 to L4: industry agents have achieved great success in fields like software engineering~\cite{tufano2024autodevautomatedaidrivendevelopment}, database interaction~\cite{pourreza2023dinsqldecomposedincontextlearning}, and web browsing~\cite{he2024webvoyagerbuildingendtoendweb}. However, progress in physical \cite{jadhav2025largelanguagemodelagent} or social domains \cite{han2024regulatormanufactureraiagentsmodeling} has been relatively slow. The essence of this difference lies in the disparity between knowledge and experience. In digitally native fields, "physical laws" are defined by APIs, code libraries, and explicit interaction protocols. The experience of an agent can be efficiently acquired through vast amounts of digital text, code, and interaction logs. However, the operational logic in physical and social fields is often filled with tacit knowledge. For example, an experienced engineer's intuition about equipment malfunction, a doctor adjusting a diagnosis based on patient emotions, or a diplomat's judgment at the negotiation table—these experiences are deeply embedded in human practice and interaction, and cannot be fully described in language or captured in data. Feedback in these domains is often delayed, ambiguous, and costly in terms of trial and error. The success of current agents largely stems from their ability to translate unstructured human knowledge into structured digital instructions. However, when the knowledge itself is unstructured, contextual, or even incommunicable, this translation paradigm fails. Therefore, the core challenge for the future is whether we should focus on using more powerful models to extract all available data to simulate tacit knowledge, or whether we need to develop new agent architectures that can collaboratively learn with human experts through a few high-quality interactions to efficiently learn experiences \cite{lim2024largelanguagemodelenabledmultiagent}.

\subsection{The Importance of Simulation Environments}

The most mature applications of industry agents are primarily concentrated in digital fields such as software engineering, data analysis, and information retrieval. In these domains, the rules of operation are typically clearly defined and formalized: code has strict syntax, APIs have clear documentation, and web pages follow standardized DOM structures. This means agents can learn and execute tasks in environments where the rules are explicit, feedback is immediate, and the cost of trial and error is low. A code interpreter, a browser DOM environment, or an API interface is itself a 100\% high-fidelity simulator of the corresponding world. In this lossless, rule-defined digital world, agents can learn and evolve through massive, low-cost interactions. Thus, the intelligence of an agent does not solely arise from the static reasoning abilities of the LLM, but rather emerges in the closed-loop interaction of "thinking-action-observation." The environment, as the key element providing "observation" and feedback, is an indispensable part of the agent's cognitive loop. Without the dynamic response of the environment, the agent's "action" loses its meaning, and its "thinking" becomes baseless. Even an embodied robot agent, trained in the most advanced physical simulators, faces significant challenges when entering the real world, where the simulator cannot fully model air resistance, ground friction, lighting changes, and sensor noise \cite{Toth2024SimtoReal}. This vast simulation-reality gap results in a sharp decline in performance \cite{Jonnarth_2025}. Therefore, the upper limit of an agent's capabilities largely depends on the quality of the environment it can interact with. In conclusion, future breakthroughs will not only rely on larger and more powerful LLMs but also on the progress in simulation engineering—whether we can create sufficiently realistic and scalable digital twin \cite{lauerschmaltz2024humandigitaltwindefinition} environments for complex physical and social systems such as manufacturing, healthcare, and finance \cite{Liu_2025}. This makes the ability to build and leverage simulation environments the fundamental prerequisite for measuring whether an industry can successfully apply advanced agents.

\subsection{Asymmetry Between Capabilities and Tasks}

When examining various agents, we encounter an interesting paradox: many systems that show significant shortcomings in core capabilities (e.g., long-term memory, complex planning) still perform excellently in specific tasks (e.g., a web scraping agent that only relies on short-term context). However, in other scenarios, even a small deficiency in capability can lead to catastrophic failure of the entire task. This reveals the asymmetric relationship between capabilities and tasks, which lies at the core of distinguishing specialists from generalists. The "wooden barrel theory" fails when tasks are highly simplified and constrained. For example, an L2-level interactive question-answering robot, which focuses on retrieval and understanding, has almost no need for long-term memory or complex planning. Here, the task boundaries limit the agent's exposed capabilities, hiding its weaknesses. On the other hand, the "wooden barrel theory" holds when tasks are open, dynamic, and long-term. An L3-level autonomous software engineer not only needs to generate code but also must understand cross-file dependencies, plan development steps, and reflect on and correct errors when they occur. Any missing step can lead to project failure. This asymmetry presents multiple choices for industry practice: should we invest resources to fill in the agent's shortcomings and create a balanced jack-of-all-trades, or should we focus on reducing the complexity of real-world problems and breaking them down into sub-tasks that can be performed by currently limited agents \cite{harper2024autogenesisagentselfgeneratingmultiagentsystems, niu2025flowmodularizedagenticworkflow}? From a practical perspective, for the foreseeable future, there will be two parallel paths: collaborative systems integrating a large number of specialized agents \cite{han2025llmmultiagentsystemschallenges}, and generalist autonomous systems that can independently handle complexity \cite{applis2025unifiedsoftwareengineeringagent, zhang2025aflowautomatingagenticworkflow}.

\subsection{The Prisoner's Dilemma of Autonomous Evolution}

From L3's self-correction to L5's autonomous goal generation, the ultimate ideal for agents is "autonomous evolution"—constantly learning, adapting, and emerging new capabilities through continuous interaction with the environment. However, this autonomy itself contains a profound contradiction, forming a prisoner's dilemma: on one hand, we desire to allow agents to explore in open environments to achieve unexpected breakthroughs (cooperative rewards) \cite{niu2025flowmodularizedagenticworkflow}; on the other hand, we fear total loss of control, concerned that they may evolve harmful or incomprehensible behaviors (the risk of betrayal). This leads to an ultimate dilemma about control and creation: how can we build a framework that allows agents to evolve autonomously while ensuring they always explore within safe boundaries \cite{sha2025agentsafetyalignmentreinforcement}? This dilemma manifests on multiple levels: (1) Goal drift—Can an agent whose initial goal is to improve productivity, during autonomous evolution, misinterpret this as reducing costs at all costs, potentially leading to safety incidents or ethical issues \cite{he2024emergedsecurityprivacyllm, berdoz2024aiagentsafelyrun}? (2) Sandbox paradox—True evolution requires interaction with real, complex environments, but for safety reasons, we can only confine agents to a controlled sandbox. Can the capabilities evolved within the sandbox effectively generalize to the real world? What are the graduation criteria from the sandbox? (3) Value locking—How can we inject a set of core values into the system at the design stage that remains robust and virtuous as the environment evolves \cite{osogami2025aiagentsregulatedbased}? Solving this dilemma may require going beyond the traditional instruction-execution paradigm, exploring new constraint mechanisms such as Constitutional AI~\cite{bai2022constitutionalaiharmlessnessai} in the agent domain, designing agent architectures capable of trustworthy self-supervision and risk assessment, and establishing a dynamic, interactive human-machine collaborative governance system. This requires us not only to design the agents themselves but also to design the social ecosystem in which they operate.

\subsection{Organizational and Process Integration Resistance}

Moving industry agents from technology validation to large-scale application often presents challenges that go beyond technology and are more rooted in organizational and process aspects. Existing IT ecosystems in enterprises often consist of legacy systems lacking modern APIs, proprietary software, and data silos, presenting significant connectivity barriers to seamless integration of agents. More importantly, the introduction of agents represents an organizational transformation, requiring employees to shift from traditional executors to managers and collaborators with agents. This inevitably encounters resistance in terms of trust-building and skill reshaping \cite{Schnur2025AgenticAI}. Potential solutions include developing low-code platforms as system connectors, establishing unified data governance platforms, and designing training and management systems for human-machine collaboration \cite{AI2027Forecast}. However, the core challenge lies in driving change management. While technical solutions can be designed, overcoming departmental silos, breaking down data silos, and reshaping employee roles and performance evaluation systems is a slow, costly, and internally competitive social process. Therefore, the successful deployment of agents depends not only on their technological advancement but also on whether industries and enterprises have the determination and capability to drive profound organizational change.

\section{Conclusion}
In this work, we present a systematic survey of LLM-driven industry agents, bridging the latest advances in their core technologies, practical applications, and evaluation methodologies. We introduce a five-level capability maturity framework to dissect how key technologies like memory, planning, and tool use evolve to support agents' progression from simple automation to complex autonomy. Our analysis reveals that current successes are predominantly confined to digital-native environments, highlighting a critical "sim-to-real gap" and a fundamental disconnect between existing evaluation metrics and the industry's demand for reliability. Based on this holistic perspective, we expect the future development of industry agents to pivot towards enhancing reliability, specialization, and human-agent synergy. By integrating advanced AI with deep domain knowledge, we believe these trustworthy agents will ultimately become a core engine of the next industrial revolution, profoundly augmenting societal productivity and creativity.


%

\appendices







\bibliographystyle{IEEEtran}
\bibliography{IEEEexample}

\end{document}